\pdfoutput=1
\documentclass[journal]{IEEEtran}

\usepackage[english]{babel}
\usepackage{xcolor}
\usepackage[table]{xcolor}
\usepackage{tabulary,booktabs,multirow}
\usepackage{listings}
\usepackage{hyperref}
\usepackage[font=small,labelfont=bf,skip=16pt]{caption}
\usepackage{subcaption}
\usepackage{tikz,float}
\usepackage[ruled,vlined]{algorithm2e}
\usepackage{orcidlink}
\usetikzlibrary{bayesnet,arrows}
\usetikzlibrary{shapes.geometric,positioning,arrows.meta,calc}
\tikzset{>=latex}
\usepackage{amsmath,amssymb,bm}
\usepackage{algorithm2e} 
\usepackage{cleveref}
\usepackage{cuted}
\usepackage{xcolor}
\usetikzlibrary{backgrounds}
\usepackage{stfloats}
\usepackage{cuted}
\usepackage[printonlyused,withpage]{acronym}

\DeclareMathOperator{\E}{\mathbb{E}}

\DeclareMathOperator*{\argmax}{arg\,max}

\DeclareMathAlphabet{\mathcal}{OMS}{cmsy}{m}{n}

\usepackage{ntheorem}
\theoremseparator{:}

\newif\ifshowhighlights\showhighlightstrue
\newcommand{\hlight}[2]{\ifshowhighlights\textcolor{#1}{#2}\else#2\fi}

\definecolor{revblue}{RGB}{0,0,190}
\newif\ifshowrev\showrevfalse
\newcommand{\rev}[1]{\ifshowrev\hlight{revblue}{#1}\else#1\fi}

\definecolor{irevgreen}{RGB}{0,110,45}
\newif\ifshowirev\showirevfalse
\newcommand{\irev}[1]{\ifshowirev\hlight{irevgreen}{#1}\else#1\fi}

\definecolor{rrevred}{RGB}{200,0,0}
\newif\ifshowrrev\showrrevfalse
\newcommand{\rrev}[1]{\ifshowrrev\hlight{rrevred}{#1}\else#1\fi}

\definecolor{ralblue}{RGB}{0,70,200}
\newif\ifshowral\showralfalse
\newcommand{\ral}[1]{\ifshowral\hlight{ralblue}{#1}\else#1\fi}

\definecolor{respred}{RGB}{200,0,0}
\newif\ifshowresp\showrespfalse
\newcommand{\resp}[1]{\ifshowresp\hlight{respred}{#1}\else#1\fi}

\newif\ifwithappendix\withappendixtrue

\newif\ifsplitrefs\splitrefstrue

\newif\ifshowauthors\showauthorstrue

\lstdefinestyle{pythonstyle}{
    language=Python,
    basicstyle=\ttfamily\small,
    breaklines=false,
    commentstyle=\color{green!50!black},
    keywordstyle=\color{blue},
    stringstyle=\color{red},
    numbers=left,
    numberstyle=\tiny\color{grey},
    numbersep=5pt,
    frame=single,
    framesep=15pt,
    rulecolor=\color{black!30},
    showstringspaces=false,
    tabsize=4
}
\lstdefinestyle{console} {
	language=bash,
	basicstyle=\footnotesize\ttfamily\singlespacing,
	backgroundcolor=\color[RGB]{248, 247, 242},
	upquote=true,
	frame=single,
	framerule=0pt,
	breakindent=0pt,
	breaklines=true,
	aboveskip=0pt,
	belowskip=4pt,
	xleftmargin=10pt,
	xrightmargin=10pt
}

\title{\texorpdfstring{\rev{LaGarNet: Latent World Models for Quasi-Static Pick-and-Place Garment Flattening}}{LaGarNet: Latent World Models for Quasi-Static Pick-and-Place Garment Flattening}}

\ifshowauthors
\author{
Halid Abdulrahim Kadi$^{1,\irev{2}}$,~\IEEEmembership{Member,~IEEE},~\orcidlink{0000-0001-9290-467X},
John Oyekan$^{1,\irev{3}}$,~\IEEEmembership{Member,~IEEE},~\orcidlink{0000-0001-6578-9928}
, and
Kasim Terzi{\'c}$^{2}$,~\IEEEmembership{Member,~IEEE},~\orcidlink{0000-0001-6692-209X}

\thanks{$^{1}$~Department of Computer Science, University of York.}
\thanks{$^{2}$~School of Computer Science, University of St Andrews.}
\thanks{$^{3}$~\irev{Loughborough University.}}

\thanks{This work is supported by a scholarship from the School of Computer Science at the University of St Andrews and by the Engineering and Physical Sciences Research Council (EPSRC) under the DIGI-CORTEX project.}}
\else
\author{Anonymous Authors}
\fi

\begin{document}

\maketitle

\begin{strip}
    \centering
    \captionsetup{type=figure}%
    \vspace{-13mm}%
    
    \subfloat[\rev{Simulated garments' initial and goal states.}\label{fig:sim-train-init}]{
        \begin{minipage}[c]{0.300\textwidth}
            \centering
\includegraphics[width=1.0\linewidth]{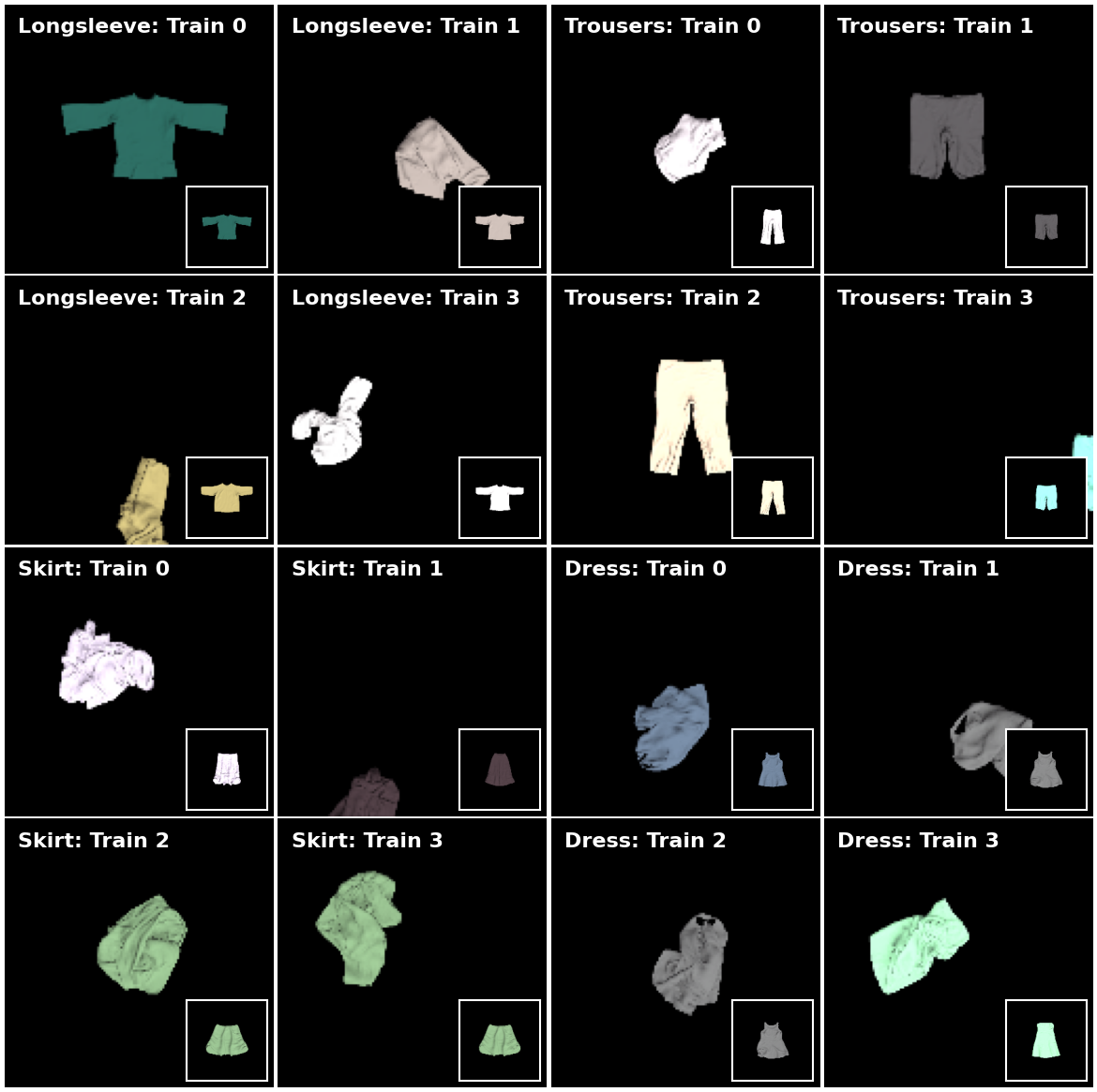}\hfill
        \end{minipage}
    }
    \subfloat[\rev{Garments used for real-world evaluation.} \label{fig:real-garments}]{
        \begin{minipage}[c]{0.415\textwidth}
            \centering
            \includegraphics[width=1.0\linewidth]{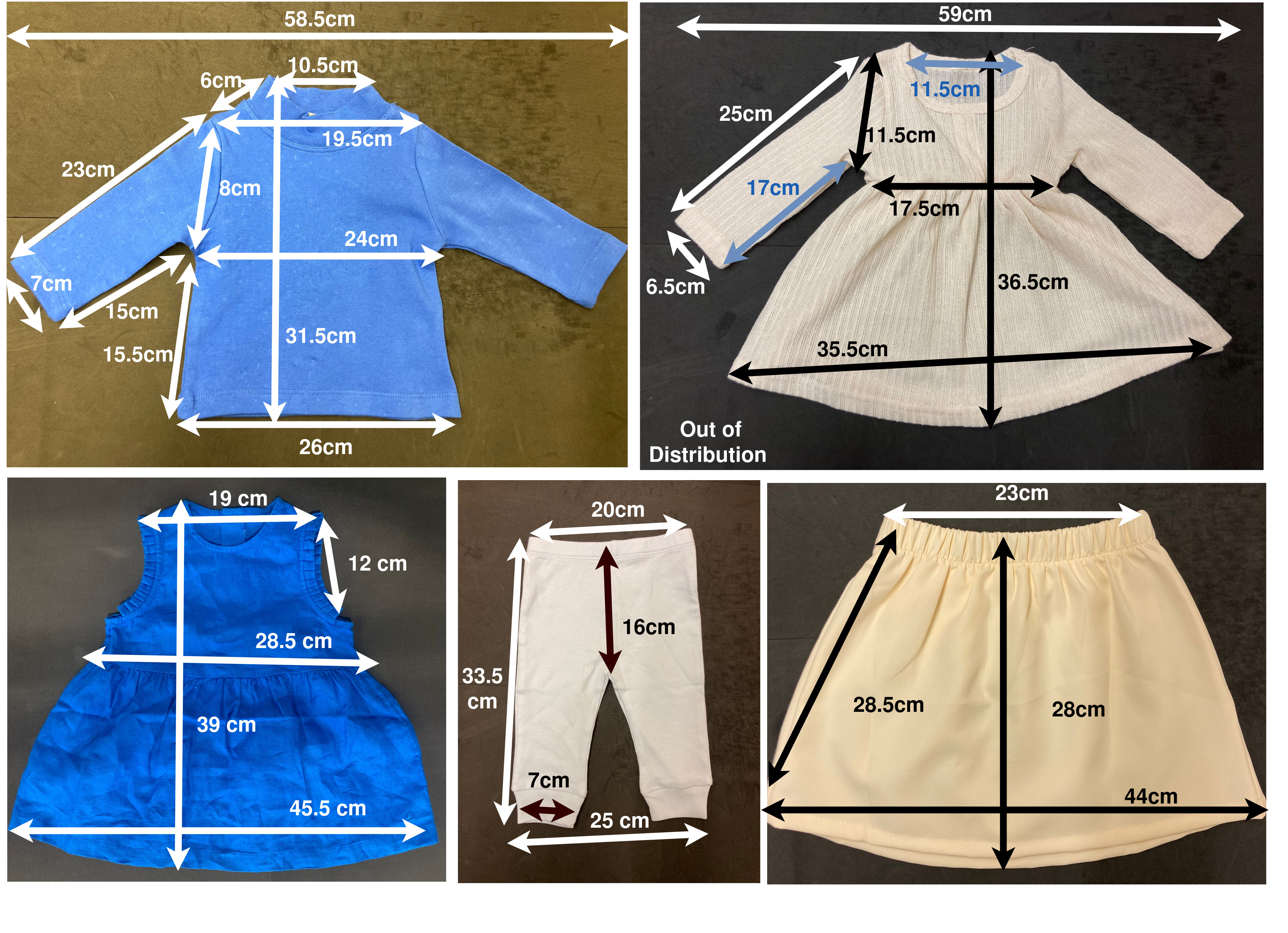}
        \end{minipage}
    }
    \subfloat[\rev{UR5e flattening a skirt.} \label{fig:ur5e-onrobot}]{
        \begin{minipage}[c]{0.222\textwidth}
            \centering
            \includegraphics[width=1.0\linewidth]{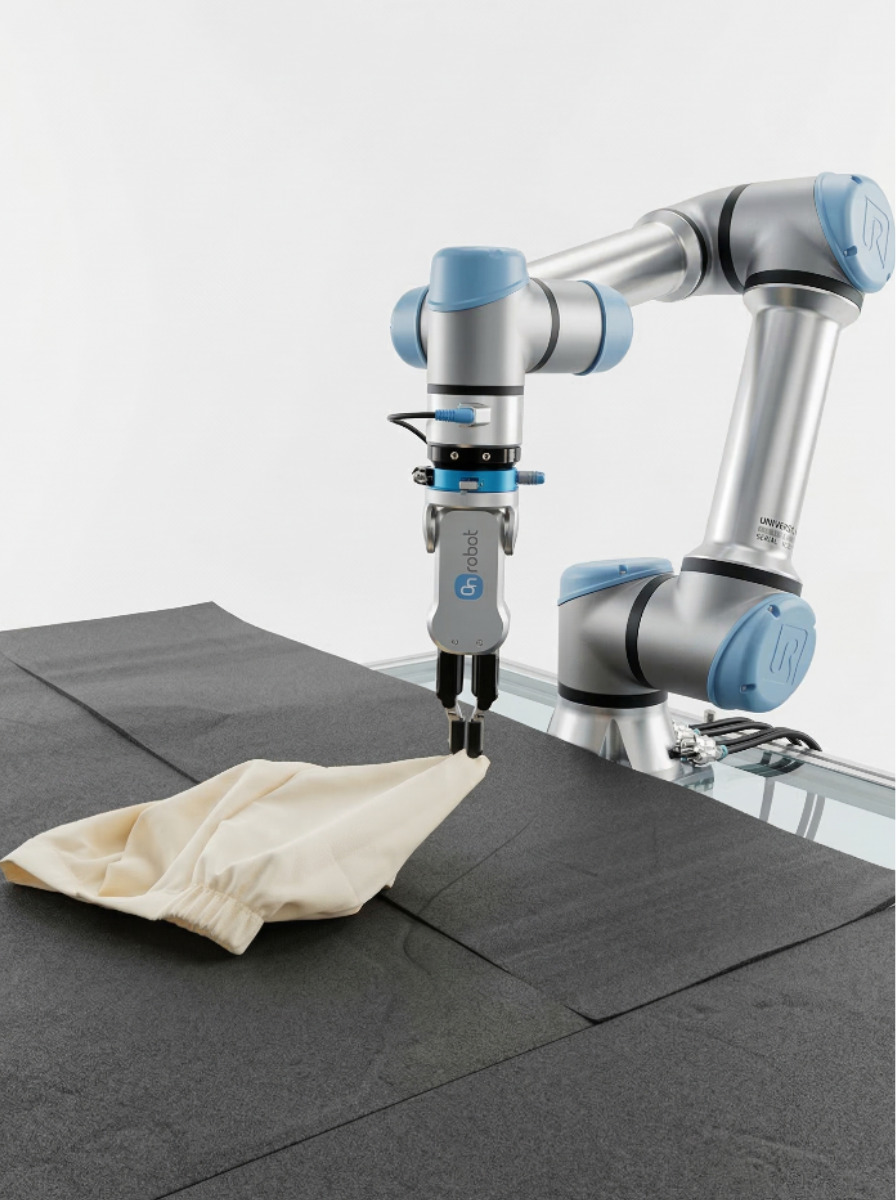}
        \end{minipage}
    }
    
    \vspace{0.12cm}
    
    \subfloat[A qualitative demonstration of LaGarNet flattening a real dress.  \label{fig:ur3e-real-action-inference}]{
        \begin{minipage} [c]{0.985\textwidth}
            \centering
            \includegraphics[width=1.0\linewidth]{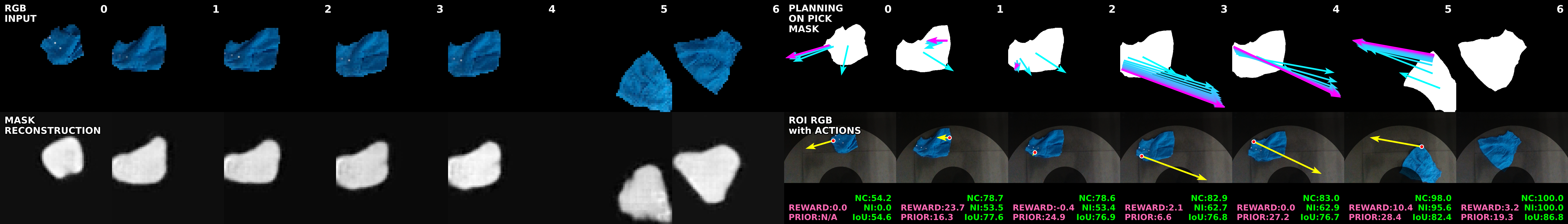}
        \end{minipage}
    }
    
    \caption{Overview of LaGarNet. \rev{\rrev{\textbf{(a)}} Initial and goal states of simulated \text{Cloth3D} garments \cite{bertiche2020cloth3d} in our benchmark. \rrev{\textbf{(b)}} Real garments used for physical evaluation, \irev{where the shape and material of the cream longsleeved dress (marked \emph{Out of Distribution}) are unseen during training}. \rrev{\textbf{(c)} Our single-arm real-world setup with a parallel gripper; input images are taken by a top-down camera.} \rrev{\textbf{(d)}} LaGarNet flattening a real dress\rrev{: blue arrows are early Model Predictive Control action means, purple the final ones (top); the bottom row reports Normalised Coverage (NC), Normalised Improvement (NI), Maximum Intersection over Union (Max IoU), and the true and predicted prior rewards ($\times$100)}. Trained in simulation, our single-policy LaGarNet transfers zero-shot to the real world for flattening 4 different types of garments.}}
    \label{fig:LaGarNet}
\end{strip}

\begin{abstract} \rev{Manipulating cloth is a grand challenge in robotics: its many degrees of freedom yield an enormous state-action space that makes learning-based control extremely difficult to train. We present a goal-conditioned recurrent state-space model (GC-RSSM) that learns the latent dynamics of pick-and-place garment manipulation. Unlike the velocity- and position-controlled settings where world models are usually studied, high-level pick-and-place is quasi-static: \resp{the garment rests between primitive actions, so consecutive observations are not temporally continuous and each transition is one large deformation.} We show that a latent world model nonetheless learns useful dynamics in this regime; our method LaGarNet matches the state-of-the-art performance of mesh-based continuous planning methods on the difficult task of garment flattening whilst using five times fewer parameters and inferring over an order of magnitude faster. \resp{Trained on our proposed coverage-alignment reward and generalisable data-collection procedure, LaGarNet is, to our knowledge, the first integrated latent world model system to flatten four garment topologies with a single policy in the real world.}} Project website: \url{https://st-cortex-lab.github.io/projects/lagarnet/}
\end{abstract}

\begin{IEEEkeywords}
Robot Learning, Cloth Manipulation, World Model, Model-Based Reinforcement Learning
\end{IEEEkeywords}

\section{Introduction} \label{sec:introduction}

Robotic manipulation of cloth-like deformable objects is a significant technical challenge with profound societal and economic implications. \rev{It is essential \resp{for garment manufacturing, domestic laundry, assistive care and sterile clinical handling, and it remains difficult because of cloth's complex, underactuated dynamics.} Garment flattening \irev{in particular is a} foundational milestone preceding tasks such as folding, ironing, and pressing.}

\rev{Two families of data-driven approaches dominate garment flattening. \textbf{Behaviour cloning (BC)} is conceptually simple, but the space of crumpled configurations is vast\rrev{,} and covering it demands an impractical volume of human demonstrations \cite{weng2022fabricflownet, lips2024learning, huang2025sis}; BC flatteners therefore \rrev{require} scripted oracles\rrev{---which are almost impossible to engineer for garments---}or narrow their scope to a single garment type. \textbf{Reinforcement learning (RL)} explores that distribution directly in simulation, enabling the most general flattening controllers \cite{matas2018sim, lin2020softgym, canberk2023cloth, gu2023learning}, almost always with \textit{model-free} algorithms. \textit{Model-based RL}, which learns and exploits a world model of the environment \cite{hafner2019learning}, remains rare on garments: cloth's large degrees of freedom make such controllers extremely difficult to train, and effective world models have stayed elusive because of deformation and self-occlusion \cite{yan2021learning, lin2020softgym}.}

\rev{Hence, we investigate whether latent world models can learn useful dynamics in quasi-static pick-and-place (PnP) garment flattening by adopting a top-down camera view and a single-gripper primitive. \resp{Contemporary behaviour-cloning and vision-language-action policies with dual-arm continuous setups struggle with long horizons and with the multi-modal action distributions of garment manipulation; a primitive compresses a whole reach-grasp-lift-place motion into four numbers, potentially reducing the state-action space that a policy needs to learn.} Whilst dual-arm multi-primitive control flattens more effectively \cite{avigal2022speedfolding}, a single arm keeps the study focused on the representation learning of the world model \cite{canberk2023cloth}. In addition, our human-in-the-loop experiments show operators flatten diverse garments exceptionally well with single-arm commands alone, indicating the primary bottleneck is \irev{reasoning about cloth deformation} rather than the hardware or the action space\resp{; our LaGarNet's latent dynamics model predicts and reasons about future observations of garment deformation and chooses which of the sampled action candidates to execute.}}

We present LaGarNet, a general single-policy method for flattening complex garments (Figure \ref{fig:LaGarNet}). \rev{It mitigates the strong inductive biases of previous data-driven cloth manipulation methods \cite{kadi2025draper, weng2022fabricflownet, hoque2022visuospatial}, which rely on domain-specific architectures, reward functions, and datasets from a particular blend of expert and random oracle policies. LaGarNet instead integrates a \irev{novel} goal-conditioned recurrent state-space model (GC-RSSM), a coverage-alignment reward and a general data-collection procedure into one system, to answer our central research question: \textit{can latent world models built for continuous control still function in quasi-static primitive setups, where consecutive observations are not temporally continuous?}}

\rrev{Planning in a compact latent space, LaGarNet} \resp{is competitive with human operators under the same single-arm constraint, although humans succeed more often at long horizons.} \irev{In a zero-shot transfer to the real world}, it averages 90\% NC, 80\% NI and 81\% Max IoU on the real garments of Figure \ref{fig:real-garments}. Our contributions are:

1. We propose a goal-conditioned recurrent state-space model (GC-RSSM) that conditions the latent inference on the goal, making both prior and posterior latent states goal-aware.

2. We develop a data-collection strategy for offline PnP garment flattening, mixing a Diffusion Policy expert learned from few demonstrations with a mask-biased random policy. \resp{It replaces the scripted oracles offline RL usually depends on, supplying the action diversity and the near-success transitions a random policy rarely reaches.}

3. We design a coverage-alignment reward for garment flattening, computable in both simulation and the real world.

\resp{Together these three components lift success rate on the simulated longsleeve benchmark within 20 actions from 6.7\% for model-free RL, 23.3\% for the closest RSSM planner and 43.3\% for behaviour cloning to 76.7\%, past the mesh-based MEDOR's 70.0\% at 17$\times$ its inference speed.}

\section{Related Work} \label{sec:related-work}

\ral{Learning-based cloth flattening and folding span behaviour cloning \cite{lips2024learning, weng2022fabricflownet, huang2025sis, chi2023diffusionpolicy}, model-free reinforcement learning (RL) \cite{matas2018sim, lin2020softgym, canberk2023cloth, gu2023learning, avigal2022speedfolding} and model-based RL \cite{yan2021learning, hoque2022visuospatial, lin2022learning, huang2022mesh}. Most successful controllers employ an explicit intermediate representation to mitigate self-occlusion, such as key points \cite{lips2024learning}, seam features \cite{huang2025sis}, a reconstructed mesh \cite{huang2022mesh} or optical flow \cite{weng2022fabricflownet}. We instead learn a latent dynamics model and obtain actions by planning, so a single common representation covers several garment types. These systems also differ in action space, from low-level position control \cite{matas2018sim, lin2020softgym} to pick-and-fling, pick-and-place (PnP) and pick-and-drag primitives, hybrids of which are highly effective \cite{avigal2022speedfolding, canberk2023cloth}. We deliberately keep a single-arm PnP setup, isolating our research question from the design choices a hybrid dual-arm action space introduces, and therefore neither evaluate nor compare against such setups.}

\ral{Model-based methods for cloth are either mesh-based or mesh-free. Visible Connectivity Dynamics \cite{lin2022learning} reconstructs a mesh of the visible cloth for more precise planning, and MEDOR \cite{huang2022mesh} extends this by recovering the occluded part from a top-down point cloud through test-time mesh fine-tuning against a garment template. Mesh-free planners such as Contrastive Forward Modelling \cite{yan2021learning} and Visual Foresight Modelling \cite{hoque2022visuospatial} learn latent or video-prediction dynamics instead, but show limited efficacy on garments and have not matched mesh-based VCD. LaGarNet is, to our knowledge, the first mesh-free planner to match mesh-based performance on garment flattening.}

\ral{The latent dynamics model PlaNet \cite{hafner2019learning}, built on a recurrent state-space model (RSSM), is widely studied in continuous control, but its reconstructions blur the edge and corner detail that shaping needs \cite{yan2021learning, lin2022learning}. \resp{PlaNet-ClothPick \cite{kadi2024planet} mitigates this with mask-restricted picks, a flattening-specific reward and an engineered offline dataset, but those same inductive biases stop it generalising beyond towels \cite{kadi2025draper}.}} \ral{Previous work has goal-conditioned RSSMs: Pertsch et al.~\cite{pertsch2020long} condition a hierarchical predictor on subtask endpoints, and Duan et al.~\cite{duan2024learning} condition actor-critic networks rather than the world model. We instead condition the latent inference itself, making both prior and posterior states goal-aware---Section \ref{sec:abl} isolates that choice against other goal-conditioning variants.}

\ral{Reward engineering is essential to the success of RL controllers, and most garment-flattening systems come with their own reward function. ClothFunnels \cite{canberk2023cloth} interpolates a normalised delta deformation distance with a delta rigid distance, Learning2Unfold \cite{gu2023learning} mixes normalised coverage with an orientation term, and SpeedFolding \cite{avigal2022speedfolding} mixes delta coverage with a learned smoothness estimate. Our coverage-alignment reward combines coverage with the IoU difference between consecutive frames and requires only top-down segmentation masks; it is therefore computable identically in simulation and reality. The effectiveness of RL also depends on the diversity of the training trajectories \cite{duan2024learning}. Prior work draws trajectories from demonstrations \cite{matas2018sim}, from corner-biased sampling that steers picks towards fabric corners and edges \cite{hoque2022visuospatial, weng2022fabricflownet}, or from other engineered strategies \cite{lin2022learning}; each imposes a strong inductive bias. We instead mix a Diffusion Policy \cite{chi2023diffusionpolicy} learned from a few human demonstrations with a mask-biased random oracle for a more generalisable collection procedure.}

\section{LaGarNet} \label{sec:planet-clothpick++}

LaGarNet is an offline model-based deep reinforcement learning approach to garment flattening, trained on simulation data with domain randomisation and transferred to real garments without any real-world fine-tuning. \ral{Its goal-conditioned recurrent state-space model (GC-RSSM) maintains a belief state over the partially observed garment configuration and scores how far each candidate action advances the task towards its goal \cite{hafner2019learning}. Trained on a coverage-alignment reward and a unified data-collection procedure, it reaches state-of-the-art flattening performance from 100,000 offline transitions.}

\subsection{Goal-Conditioned Recurrent State-Space Model} \label{sec:gc-rssm}

\rev{LaGarNet builds on State-Space Models (SSMs), generative frameworks for sequential data that support inferring future states and learning representations. Hafner et al.~\cite{hafner2019learning} define the Recurrent State-Space Model (RSSM) under the Partially Observable Markov Decision Process (POMDP) formulation, with latent state dynamics:}
\begin{equation}
\begin{aligned}
    \text{Recurrent dynamic model:} & \quad \bm{h}_t=f(\bm{h}_{t-1},\bm{z}_{t-1},\bm{a}_{t-1}) \\
    \text{Representation model:}    & \quad \hat{\bm{z}}_t\sim q(\hat{\bm{z}}\;|\;\bm{h}_t,\bm{x}_t) \\
    \text{Transition predictor:}    & \quad \tilde{\bm{z}}_t\sim p(\tilde{\bm{z}}\;|\;\bm{h}_{t}),
\end{aligned}
\end{equation}
where $\bm{x}$ represents the observation, $\bm{a}$ represents the action, $\bm{h}$ represents the deterministic latent representation, and ($\hat{\bm{z}}$, $\tilde{\bm{z}}$) represent the posterior and prior stochastic latent states. \rev{\rrev{In LaGarNet, the deterministic state $\bm{h}$ and the stochastic state $\bm{z}$ have 300 and 60 dimensions, respectively.}}

\rev{LaGarNet trains a goal-conditioned latent dynamics model by learning a feature representation $\bm{y}$, extending the RSSM with goal-aware inference and generation (Figure~\ref{fig:gc-rssm-pgm}): the goal observation $\bm{g}$ enters the estimation of the stochastic latent state $\bm{z}$, \irev{defining} the GC-RSSM as:}
\begin{align}
    \hat{\bm{z}}_t &\sim q(\hat{\bm{z}} \;|\; \bm{h}_t, \bm{x}_t, \bm{g}) 
    && \text{GC Representation Model} \label{eq:gc-repr} \\
    \tilde{\bm{z}}_t &\sim p(\tilde{\bm{z}} \;|\; \bm{h}_t, \bm{g}) 
    && \text{GC Transition Predictor} \label{eq:gc-trans}
\end{align}

\ral{As shown in Figure \ref{fig:network}, LaGarNet comprises four components with ReLU activations throughout: a convolutional image encoder, a transposed-convolutional decoder, a goal-conditioned transition model built from MLPs and a Gated Recurrent Unit, and a 5-layer MLP reward predictor.} Training maximises the marginal log-likelihood of the estimated features and rewards given the action sequence and the goal observation, $\argmax_{\theta} \ln p_{\theta}(\bm{y}_{1:T}, r_{1:T} | \bm{a}_{1:T}, \bm{g})$. \ral{Treating this as variational inference over the factorised generative model $\prod_t p(\bm{y}_t|\bm{z}_t) p(r_t|\bm{z}_t) p(\bm{z}_t|\bm{z}_{t-1}, \bm{a}_{t-1}, \bm{g})$ and posterior $\prod_t q(\bm{z}_t| \bm{x}_{1:t}, \bm{a}_{1:t-1}, \bm{g})$ (in which the deterministic states $\bm{h}_t$ are abstracted because they follow deterministically from previous states) gives the GC-RSSM evidence lower bound:}
\begin{align}
    \mathcal{L} =& \sum_{t=1}^{T} \mathbb{E}_{q(\bm{z}_{1:t-1})} \Bigg[ - 
     \underset{\bm{z}_t \sim q}{\E} \Big[ \ln p(\bm{y}_t|\bm{z}_t) + \ln p(r_t|\bm{z}_t) \Big] \nonumber\\
    &+ \text{KL}\Big[ q(\bm{z}_t|\bm{x}_{1:t}, \bm{a}_{1:{t-1}}, \bm{g}) \;\Big|\Big|\; p(\bm{z}_{t}|\bm{z}_{t-1}, \bm{a}_{t-1}, \bm{g}) \Big]
     \Bigg] \label{eqn:planet-objective-1}.
\end{align}

\ral{The $64 \times 64$ goal image $\bm{g}$ passes through the same convolutional encoder as the current observation to give the 1024-dimensional embedding $\bm{e}_g$. Then, we use the reparameterisation trick to sample $\bm{z}$, i.e., $\bm{z} = \bm{\mu} + \bm{\sigma} \odot \bm{\epsilon}$ with $\bm{\epsilon} \sim \mathcal{N}(\bm{0}, \bm{I})$, its standard deviation kept positive by a softplus plus a floor $\sigma_{\text{min}} = 0.1$.} We use MSE losses for feature and reward prediction, equivalent to the negative log-likelihood under a Gaussian and weighted by $\gamma_y = \gamma_r = 1.0$ in Equation \ref{eqn:total-loss}. \ral{For latent regularisation we adopt KL balancing, which scales prior and posterior gradients differently so the prior moves towards the posterior faster than the reverse, with a ``free nat'' threshold $c = 1$ against posterior collapse \cite{hafner2019learning}:}
\begin{equation}
\begin{aligned}
    \mathcal{L}_{KL} = &\beta_{KL} \max\Big(c, \text{KL}\big[\text{sg}(q(\hat{\bm{z}}_t)) \;\big|\big|\; p(\tilde{\bm{z}}_t)\big]\Big) \\ 
    &+ (1 - \beta_{KL}) \max\Big(c, \text{KL}\big[q(\hat{\bm{z}}_t) \;\big|\big|\; \text{sg}(p(\tilde{\bm{z}}_t))\big]\Big),
\end{aligned} \label{eqn:kl-balance}
\end{equation}
\ral{where $\text{sg}(\cdot)$ is the stop-gradient operator, $\beta_{KL} = 0.8$ and $\gamma_{KL} = 1.0$. For consistent multi-step prediction, we add overshooting losses, computed from more than one-step priors, on the KL divergence and the reward predictions of future prior states across $T_f$ steps (weighted by $\gamma_{KLo} = \gamma_{ro} = 0.1$). The total loss is:}
\begin{equation}
    \mathcal{L}_{\text{total}} = \gamma_y \mathcal{L}_y + \gamma_r \mathcal{L}_r + \gamma_{KL} \mathcal{L}_{KL} + \irev{\gamma_{KLo} \mathcal{L}_{KLo}} + \gamma_{ro} \mathcal{L}_{ro}, \label{eqn:total-loss}
\end{equation}
\ral{where $\mathcal{L}_y$ and $\mathcal{L}_r$ are the feature and reward MSE losses and $\mathcal{L}_{KLo}$, $\mathcal{L}_{ro}$ the corresponding overshooting losses.}

At test time, LaGarNet takes $(\bm{x}_t, \bm{g})$ and maintains the posterior latent states $(\bm{h}_t, \hat{\bm{z}}_t)$, updating them with the previous action and the current observation; \ral{latent vectors start at zero and the initial posterior follows from a no-operation action $\bm{a}_{\text{no-op}} = \bm{1}$.}
\rev{It selects the first action of a $T_f$-step plan by Model Predictive Control with the Cross-Entropy Method (MPC-CEM):}
\begin{align}
   \bm{a}_{t:t+T_f-1} = \argmax_{\bm{a}} \sum_{k=t}^{t+T_f-1} r_\theta(\tilde{\bm{z}}_{k+1}(\bm{a}_k)),
\end{align}
where $\tilde{\bm{z}}_{k+1}(\bm{a}_k) \sim p(\tilde{\bm{z}}_{k+1} \;|\; \bm{h}_{k+1}=f(\bm{h}_k, \hat{\bm{z}}_k, \bm{a}_k), \bm{g})$. In our setup we set the prior future prediction horizon to $T_f=1$\rrev{;} \irev{Section \ref{sec:abl} shows that longer horizons degrade performance: a single primitive produces a large deformation whose prediction error compounds over multi-step rollouts}.

CEM solves this over the normalised action space $[-1,1]^4$, whose dimensions encode pick and place positions in normalised pixel coordinates. \ral{From a diagonal Gaussian initialised at $\bm{\mu}=\bm{0}$, $\bm{\sigma}=\bm{1}$ we run $K=100$ iterations, each drawing $N=5000$ candidates and rejecting, without resampling, any whose pick pixel falls outside the cloth mask or the robot's workspace mask or whose place pixel falls outside the workspace mask; the survivors are scored by $r_\theta$ and $\bm{\mu}, \bm{\sigma}$ refitted on the best $10\%$ ($N_e=500$ elites). The clipped mean of the final distribution is executed on the robot.}

\subsection{Coverage-Alignment Reward}  \label{sec:ca-reward}

\begin{figure*}[t!]
    \centering
    
    \begin{subfigure}[t]{0.67\linewidth}
        \centering
        \scalebox{0.58}{\tikz{
 \tikzset{
   half-half/.style={
     path picture={
       \fill[orange!40] (path picture bounding box.south west) rectangle (path picture bounding box.north);
       \fill[red!80] (path picture bounding box.south) rectangle (path picture bounding box.north east);
     }
   }
 }

 \node[latent, minimum size=0.8cm, fill=orange!40] (z0) {$\boldsymbol{z}_0$};
 \node[latent, rectangle, minimum size=0.8cm, above=0.6cm of z0] (h0) {$\boldsymbol{0}$};
 \node[obs, minimum size=0.8cm, above=0.6cm of h0] (a0) {$\boldsymbol{a}_{\text{no-op}}$}; 
 
 \node[latent, minimum size=0.8cm, below=0.6cm of z0, fill=blue!40] (y0) {$\boldsymbol{y}_1$};
 \node[latent, minimum size=0.8cm, below=0.25cm of y0, fill=blue!40] (r0) {$\boldsymbol{r}_1$};
 
 \node[latent, minimum size=0.8cm, right=1.8cm of z0, half-half] (z1) {$\boldsymbol{z}_1$};
 \node[latent, rectangle, minimum size=0.8cm, above=0.6cm of z1] (h1) {$\boldsymbol{h}_{1}$};
 \node[obs, minimum size=0.8cm, above=0.6cm of h1] (a1) {$\boldsymbol{a}_1$}; 
 
 \node[latent, minimum size=0.8cm, below=0.6cm of z1, fill=blue!40] (y1) {$\boldsymbol{y}_1$};
 \node[latent, minimum size=0.8cm, below=0.25cm of y1, fill=blue!40] (r1) {$\boldsymbol{r}_1$};
 
 \node[obs, minimum size=0.8cm, left=0.35cm of z1, fill=green!30] (g1) {$\boldsymbol{g}$};
 \node[obs, minimum size=0.8cm, left=0.35cm of y1, fill=purple!30] (x1) {$\boldsymbol{x}_1$};

 \node[latent, minimum size=0.8cm, right=1.8cm of z1, half-half] (z2) {$\boldsymbol{z}_2$};
 \node[latent, rectangle, minimum size=0.8cm, above=0.6cm of z2] (h2) {$\boldsymbol{h}_{2}$};
 \node[obs, minimum size=0.8cm, above=0.6cm of h2] (a2) {$\boldsymbol{a}_2$}; 
 
 \node[latent, minimum size=0.8cm, below=0.6cm of z2, fill=blue!40] (y2) {$\boldsymbol{y}_2$};
 \node[latent, minimum size=0.8cm, below=0.25cm of y2, fill=blue!40] (r2) {$\boldsymbol{r}_2$};
 
 \node[obs, minimum size=0.8cm, left=0.35cm of z2, fill=green!30] (g2) {$\boldsymbol{g}$};
 \node[obs, minimum size=0.8cm, left=0.35cm of y2, fill=purple!30] (x2) {$\boldsymbol{x}_2$};

 \node[latent, minimum size=0.8cm, right=1.8cm of z2, half-half] (zt) {$\boldsymbol{z}_{t-1}$};
 \node[latent, rectangle, minimum size=0.8cm, above=0.6cm of zt] (ht) {$\boldsymbol{h}_{t-1}$};
 \node[obs, minimum size=0.8cm, above=0.6cm of ht] (at) {$\boldsymbol{a}_{t-1}$}; 
 
 \node[latent, minimum size=0.8cm, below=0.6cm of zt, fill=blue!40] (yt) {$\boldsymbol{y}_{t-1}$};
 \node[latent, minimum size=0.8cm, below=0.25cm of yt, fill=blue!40] (rt) {$\boldsymbol{r}_{t-1}$};
 
 \node[obs, minimum size=0.8cm, left=0.35cm of zt, fill=green!30] (gt) {$\boldsymbol{g}$};
 \node[obs, minimum size=0.8cm, left=0.35cm of yt, fill=purple!30] (xt) {$\boldsymbol{x}_{t-1}$};

 \node[latent, minimum size=0.8cm, right=1.8cm of zt, half-half] (zt1) {$\boldsymbol{z}_{t}$};
 \node[latent, rectangle, minimum size=0.8cm, above=0.6cm of zt1] (ht1) {$\boldsymbol{h}_{t}$};
 \node[latent, minimum size=0.8cm, above=0.6cm of ht1, fill=yellow!30] (at1) {$\boldsymbol{a}_{t}$}; 
 
 \node[latent, minimum size=0.8cm, below=0.6cm of zt1, fill=blue!40] (yt1) {$\boldsymbol{y}_{t}$};
 \node[latent, minimum size=0.8cm, below=0.25cm of yt1, fill=blue!40] (rt1) {$\boldsymbol{r}_{t}$};
 
 \node[obs, minimum size=0.8cm, left=0.35cm of zt1, fill=green!30] (gt1) {$\boldsymbol{g}$};
 \node[obs, minimum size=0.8cm, left=0.35cm of yt1, fill=purple!30] (xt1) {$\boldsymbol{x}_{t}$};

 \node[latent, minimum size=0.8cm, right=1.8cm of zt1, fill=red!80] (zt2) {$\boldsymbol{z}_{t+1}$};
 \node[latent, rectangle, minimum size=0.8cm, above=0.6cm of zt2] (ht2) {$\boldsymbol{h}_{t+1}$};
 \node[latent, minimum size=0.8cm, above=0.6cm of ht2, fill=yellow!30] (at2) {$\boldsymbol{a}_{t+1}$}; 
 
 \node[latent, minimum size=0.8cm, below=0.6cm of zt2, fill=blue!15] (yt2) {$\boldsymbol{y}_{t+1}$};
 \node[latent, minimum size=0.8cm, below=0.25cm of yt2, fill=blue!15] (rt2) {$\boldsymbol{r}_{t+1}$};
 
 \node[obs, minimum size=0.8cm, left=0.35cm of zt2, fill=green!30] (gt2) {$\boldsymbol{g}$};

 \node[latent, minimum size=0.8cm, right=1.8cm of zt2, fill=red!80] (zT) {$\boldsymbol{z}_T$};
 \node[latent, rectangle, minimum size=0.8cm, above=0.6cm of zT] (hT) {$\boldsymbol{h}_{T}$};
 \node[latent, minimum size=0.8cm, above=0.6cm of hT, fill=yellow!30] (aT) {$\boldsymbol{a}_{T}$}; 
 
 \node[latent, minimum size=0.8cm, below=0.6cm of zT, fill=blue!15] (yT) {$\boldsymbol{y}_{T}$};
 \node[latent, minimum size=0.8cm, below=0.25cm of yT, fill=blue!15] (rT) {$\boldsymbol{r}_{T}$};
 
 \node[obs, minimum size=0.8cm, left=0.35cm of zT, fill=green!30] (gT) {$\boldsymbol{g}$};

 \edge[black, very thick] {a0, h0, z0} {h1} 
 \edge[black, very thick] {a1, h1, z1} {h2} 
 \edge[black, very thick] {at, ht, zt} {ht1} 
 \edge[black, very thick] {at1, ht1, zt1} {ht2}
 
 \edge[blue, very thick] {z0}{y0} \edge[blue, very thick] {z1}{y1} \edge[blue, very thick] {z2}{y2} 
 \edge[blue, very thick] {zt}{yt} \edge[blue, very thick] {zt1}{yt1} \edge[blue, very thick] {zt2}{yt2} \edge[blue, very thick] {zT}{yT}
 
 \path (z0) edge [bend left=45, ->, blue, very thick] (r0); \path (h0) edge [bend left=45, ->, blue, very thick] (y0); \path (h0) edge [bend left=45, ->, blue, very thick] (r0);
 \path (z1) edge [bend left=45, ->, blue, very thick] (r1); \path (h1) edge [bend left=45, ->, blue, very thick] (y1); \path (h1) edge [bend left=45, ->, blue, very thick] (r1);
 \path (z2) edge [bend left=45, ->, blue, very thick] (r2); \path (h2) edge [bend left=45, ->, blue, very thick] (y2); \path (h2) edge [bend left=45, ->, blue, very thick] (r2);
 \path (zt) edge [bend left=45, ->, blue, very thick] (rt); \path (ht) edge [bend left=45, ->, blue, very thick] (yt); \path (ht) edge [bend left=45, ->, blue, very thick] (rt);
 \path (zt1) edge [bend left=45, ->, blue, very thick] (rt1); \path (ht1) edge [bend left=45, ->, blue, very thick] (yt1); \path (ht1) edge [bend left=45, ->, blue, very thick] (rt1);
 \path (zt2) edge [bend left=45, ->, blue, very thick] (rt2); \path (ht2) edge [bend left=45, ->, blue, very thick] (yt2); \path (ht2) edge [bend left=45, ->, blue, very thick] (rt2);
 \path (zT) edge [bend left=45, ->, blue, very thick] (rT); \path (hT) edge [bend left=45, ->, blue, very thick] (yT); \path (hT) edge [bend left=45, ->, blue, very thick] (rT);

 \path (g1) edge [densely dashed, bend left=25, ->, orange, very thick] (z0); 
 \path (g1) edge [densely dashed, bend left=25, ->, orange, very thick] (z1);
 \path (g2) edge [densely dashed, bend left=25, ->, orange, very thick] (z2);
 \path (gt) edge [densely dashed, bend left=25, ->, orange, very thick] (zt);
 \path (gt1) edge [densely dashed, bend left=25, ->, orange, very thick] (zt1);

 \path (x1) edge [densely dashed, bend left=15, ->, orange, very thick] (z0); 
 \path (x1) edge [densely dashed, bend left=15, ->, orange, very thick] (z1);
 \path (x2) edge [densely dashed, bend left=15, ->, orange, very thick] (z2);
 \path (xt) edge [densely dashed, bend left=15, ->, orange, very thick] (zt);
 \path (xt1) edge [densely dashed, bend left=15, ->, orange, very thick] (zt1);

 \path (h0) edge [densely dashed, bend right=30, ->, orange, very thick] (z0);
 \path (h1) edge [densely dashed, bend right=30, ->, orange, very thick] (z1);
 \path (h2) edge [densely dashed, bend right=30, ->, orange, very thick] (z2);
 \path (ht) edge [densely dashed, bend right=30, ->, orange, very thick] (zt);
 \path (ht1) edge [densely dashed, bend right=30, ->, orange, very thick] (zt1);

 \edge[red, very thick] {h1}{z1} \edge[red, very thick] {h2}{z2} \edge[red, very thick] {ht}{zt} \edge[red, very thick] {ht1}{zt1} \edge[red, very thick] {ht2}{zt2} \edge[red, very thick] {hT}{zT}
 
 \path (g1) edge [->, red, very thick] (z1);
 \path (g2) edge [->, red, very thick] (z2);
 \path (gt) edge [->, red, very thick] (zt);
 \path (gt1) edge [->, red, very thick] (zt1);
 \path (gt2) edge [->, red, very thick] (zt2);
 \path (gT) edge [->, red, very thick] (zT);

 \path (a2) -- node {\dots} (at); \path (h2) -- node {\dots} (ht); 
 \path (r2) -- node {\dots} (rt);

 \path (at2) -- node {\dots} (aT); \path (ht2) -- node {\dots} (hT); 
 \path (yt2) -- node {\dots} (yT); \path (rt2) -- node {\dots} (rT);
}}
        \caption{Probabilistic Graphical Model of GC-RSSM}
        \label{fig:gc-rssm-pgm}
    \end{subfigure}
    \setcounter{subfigure}{2}%
    \begin{subfigure}[t]{0.32\linewidth}
        \centering
        \includegraphics[width=\linewidth]{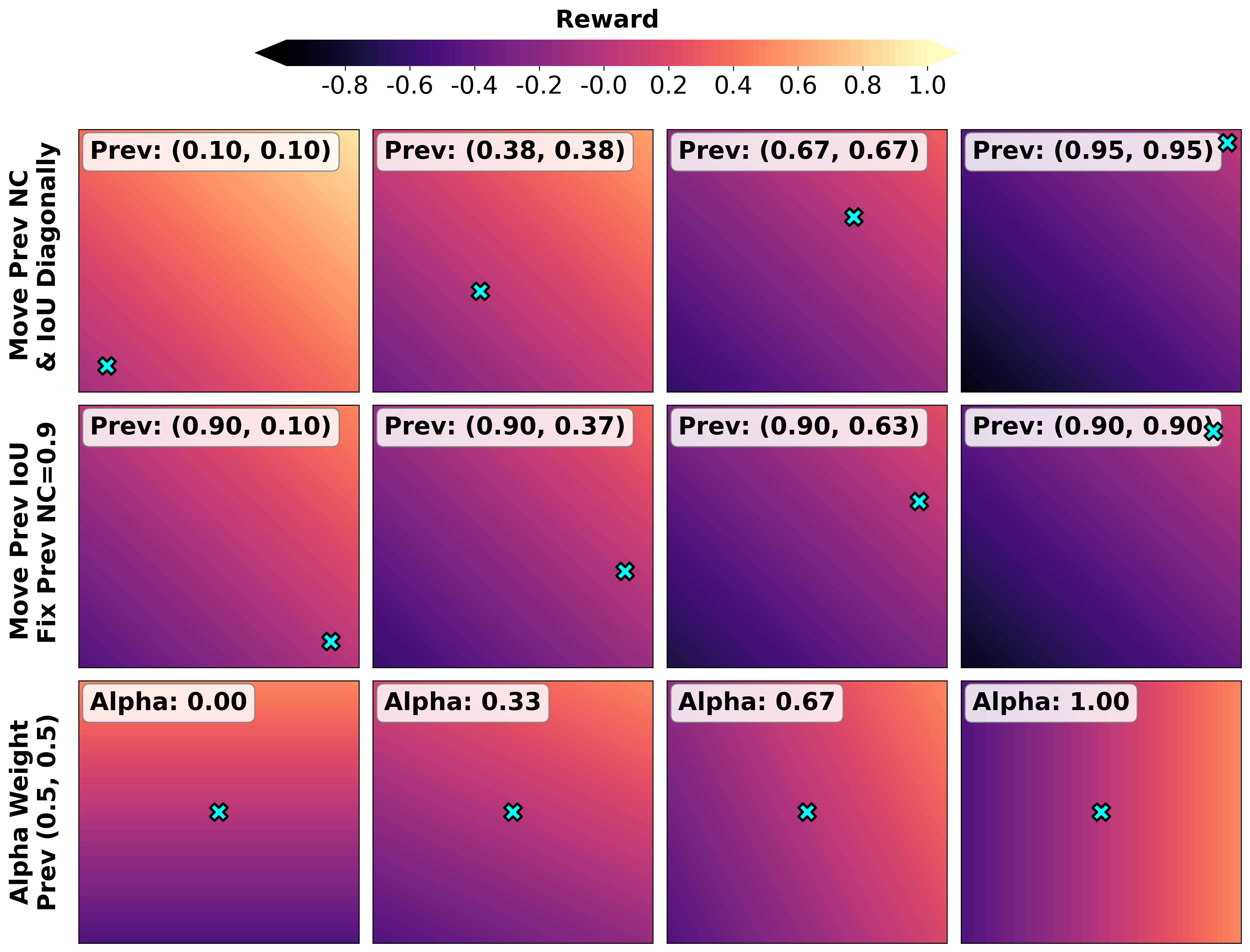}
        \caption{Coverage-Alignment Reward}
        \label{fig:reward-function}
    \end{subfigure}

    \vspace{6pt}

    \setcounter{subfigure}{1}%
    \begin{subfigure}[b]{0.67\linewidth}
        \centering
        \scalebox{0.56}{\tikz[
    >=stealth,
    block/.style={rectangle, draw, thick, align=center, minimum height=1cm, minimum width=1.5cm, rounded corners=2pt},
    mlpblock/.style={rectangle, draw, thick, align=center, minimum height=0.7cm, minimum width=1.1cm, rounded corners=2pt},
    zblock/.style={rectangle, draw, thick, align=center, minimum height=0.8cm, minimum width=1.8cm, rounded corners=2pt},
    square/.style={rectangle, draw, thick, minimum size=2cm, fill=white},
    funnel/.style={trapezium, draw, thick, trapezium stretches body, shape border rotate=270, minimum height=1.5cm, minimum width=1.2cm, fill=blue!10},
    dec_funnel/.style={trapezium, draw, thick, trapezium stretches body, shape border rotate=90, minimum height=1.5cm, minimum width=1.2cm, fill=blue!10},
    flat/.style={rectangle, draw, thick, minimum height=0.6cm, minimum width=2cm, fill=white, align=center},
    slim/.style={rectangle, draw, thick, minimum height=1.2cm, minimum width=0.4cm, fill=white, align=center},
    half split/.style={
        path picture={
            \fill[orange!40] (path picture bounding box.south west) rectangle (path picture bounding box.north);
            \fill[red!80] (path picture bounding box.south) rectangle (path picture bounding box.north east);
        }
    }
]{
    \node[block, fill=gray!60] at (0, 0) (gcrssm) {GRU};

    \node[mlpblock, fill=gray!60] at (0, 1.5) (actmlp) {MLP};

    \node[zblock, fill=orange!40] at (0, 3) (zhat) {$\boldsymbol{\hat{z}}_{1...T}$};
    \node[zblock, fill=red!80] at (0, 4.5) (ztilde) {$\boldsymbol{\tilde{z}}_{1...T}$};

    \node[mlpblock, fill=orange!40] at (-2.5, 3) (postmlp) {MLP};
    \node[mlpblock, fill=red!80] at (-2.5, 4.5) (priormlp) {MLP};

    \node[flat] at (-2.5, 0) (h) {$\boldsymbol{h}_{1...T}$};

    \node[mlpblock, fill=blue!20] at (2, 0) (jointmlp) {MLP};

    \node[flat, fill=gray!20] at (-5.7, 1.5) (action) {$\boldsymbol{a}_{\text{no-op},1...{t-1}}$};
    \node[flat, fill=yellow!20] at (-5.7, 0.9) (action_samp) {$\boldsymbol{a}_{{t...T}}$};

    \node[slim, fill=green!30, label={above:$\boldsymbol{e}^g$}] at (-5.7, 4.5) (eg_box) {};
    \node[funnel, left=0.2cm of eg_box, fill=gray!30] (cnng) {CNN};
    \node[square, left=0.2cm of cnng, fill=green!30, label={above:$\boldsymbol{g}$}] (g_box) {\includegraphics[width=1.8cm, height=1.8cm]{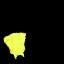}};

    \node[slim, fill=purple!30, label={above:$\boldsymbol{e}^x_{1...t}$}] at (-5.7, 2.5) (ex_box) {};
    \node[funnel, left=0.2cm of ex_box, fill=gray!30] (cnnx) {CNN};
    \node[square, left=0.2cm of cnnx, fill=purple!30, label={below:$\boldsymbol{x}_{1...t}$}] (x_box) {\includegraphics[width=1.8cm, height=1.8cm]{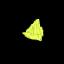}};

    \begin{scope}[on background layer]
        \node[draw=purple!50, thick, fill=none, rounded corners=5pt, fit=(zhat) (ztilde), inner sep=0.2cm] (zgroup) {};

        \coordinate (box_bottom) at (0, -1);
        \coordinate (box_left) at (-4.0, 0); 
        \node[draw=gray!80, ultra thick, dashed, fill=none, rounded corners=10pt, fit=(priormlp) (postmlp) (zgroup) (actmlp) (gcrssm) (h) (jointmlp) (box_bottom) (box_left), inner sep=0.42cm] (gcrssm_box) {};
        \node[above left, font=\bfseries, xshift=2.1cm, yshift=0cm, text=gray!80] at (ztilde.east) {GC-RSSM};
    \end{scope}

    \node[mlpblock, fill=blue!40] at (4.3, 4) (decmlp) {MLP};
    \node[dec_funnel, fill=blue!40] at (4.3, 0) (deccnn) {CNN};

    \node[flat, minimum width=1.5cm, fill=blue!40] at (6.3, 4) (rhat) {$\boldsymbol{\hat{r}}_{1...T}$};
    \node[flat, minimum width=1.5cm, fill=blue!40] at (6.3, 5.5) (rgt) {$\boldsymbol{r}_{1...T}$}; 

    \node[square, label={below:$\boldsymbol{\hat{y}}_{1...T}$}, fill=blue!40] at (6.3, 0) (yhat) {\includegraphics[width=1.8cm, height=1.8cm]{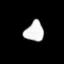}};
    \node[square, label={left:$\boldsymbol{y}_{1...T}$}, fill=blue!40] at (6.3, 2.5) (ygt) {\includegraphics[width=1.8cm, height=1.8cm]{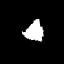}};

    \node[square, below=1.4cm of x_box, xshift=0.6cm,label={below:$\boldsymbol{x}_\text{org}$}] (x_org) {\includegraphics[width=1.8cm, height=1.8cm]{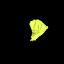}};
    \node[square, right=0.2cm of x_org,label={below:$\boldsymbol{g}_\text{org}$}] (goal_org) {\includegraphics[width=1.8cm, height=1.8cm]{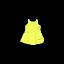}};

    \node[draw=gray!80, ultra thick, dashed, rounded corners=8pt, fit=(x_org) (goal_org), inner sep=0.4cm] (dataset_box) {};

    \draw[->, ultra thick, dashed, black, rounded corners] (dataset_box.west) -- ++(-0.6, 0) |- node[pos=0.25, right, align=left, font=\footnotesize\bfseries] {Data Augmentation\\ (training only)} (x_box.west);

    \draw[->, ultra thick, dashed, black, rounded corners] (dataset_box.west) -- ++(-0.6, 0) |-  (g_box.west);
    
    \draw[->, ultra thick, red] ([yshift=0.2cm]eg_box.east) -- ([yshift=0.2cm]priormlp.west);
    \draw[->, ultra thick, dashed, orange] ([yshift=0.3cm]ex_box.east) -- ([yshift=-0.2cm]postmlp.west);
    
    \draw[->, ultra thick, dashed, orange, rounded corners] (eg_box.east) -- ++(0.6,0) |- ([yshift=0.2cm]postmlp.west);

    \draw[->, ultra thick, red] (priormlp.east) -- node[above, yshift=2pt] {\tikz[scale=0.3]{\draw[thick] (-1.2,0) -- (1.2,0); \draw[thick, fill=red!80] (-1,0) .. controls (-0.5,0) and (-0.3,1.5) .. (0,1.5) .. controls (0.3,1.5) and (0.5,0) .. (1,0);}} (ztilde.west);
    
    \draw[->, ultra thick, dashed, orange] (postmlp.east) -- node[above, yshift=2pt] {\tikz[scale=0.3]{\draw[thick] (-1.2,0) -- (1.2,0); \draw[thick, fill=orange!40] (-1,0) .. controls (-0.5,0) and (-0.3,1.5) .. (0,1.5) .. controls (0.3,1.5) and (0.5,0) .. (1,0);}} (zhat.west);
    
    \draw[<->, ultra thick, purple, dashed] (ztilde.south) -- node[font=\bfseries] {KL Balance} (zhat.north);

    \draw[->, ultra thick, dashed, orange] (h.north) -- (postmlp.south);
    
    \draw[->, ultra thick, red, rounded corners] (h.west) -- ++(-0.6, 0) |- ([yshift=-0.2cm]priormlp.west);

    \draw[->, ultra thick, solid, black] (zhat.south) -- (actmlp.north);
    
    \draw[->, ultra thick, black, rounded corners] (action.east) -- (actmlp.west);

    \draw[->, ultra thick, black] (actmlp.south) -- (gcrssm.north);

    \draw[->, ultra thick, black] (gcrssm.west) -- (h.east);
    \draw[->, ultra thick, black, rounded corners] (h.south) -- ++(0,-0.55) -| (gcrssm.south);

    \draw[->, ultra thick, blue, rounded corners] (zgroup.east) -| (jointmlp.north);
    \draw[->, ultra thick, blue, rounded corners] ([xshift=-0.2cm]h.south) -- ++(0,-1.0) -| (jointmlp.south);

    \draw[->, ultra thick, blue, rounded corners] (jointmlp.east) -- ++(0.6,0) |- (decmlp.west);
    \draw[->, ultra thick, blue, rounded corners] (jointmlp.east) -- (deccnn.west);

    \draw[<->, ultra thick, dashed, purple] (rhat.north) -- node[right, font=\bfseries] {MSE} (rgt.south);

    \draw[<->, ultra thick, dashed, purple] (yhat.north) -- node[right, font=\bfseries] {MSE} (ygt.south);
}}
        \caption{Network Architecture and Training}
        \label{fig:network}
    \end{subfigure}
    \setcounter{subfigure}{3}%
    \begin{subfigure}[b]{0.32\linewidth}
        \centering
        \scalebox{0.74}{\begin{tikzpicture}[
            >={Stealth[length=2mm]},
            node distance=5mm,
            box/.style={rectangle, draw, thick, rounded corners=2pt, align=center,
                        inner sep=3pt, font=\footnotesize, minimum height=6.5mm},
            dec/.style={diamond, draw, thick, aspect=2, align=center,
                        inner sep=1pt, font=\footnotesize},
            lbl/.style={font=\scriptsize, inner sep=1pt}
        ]
            \node[box, fill=gray!10] (start) {Start: $\mathcal{D}\leftarrow\emptyset$};
            \node[dec, below=4mm of start, fill=gray!15] (dec) {$u\sim\mathcal{U}(0,1)$};
            \node[box, below left=9mm and 2mm of dec, fill=blue!15]  (diff) {Diffusion policy\\$\pi_{\text{diff}}$};
            \node[box, below right=9mm and 2mm of dec, fill=orange!20] (rdm) {Mask-biased random\\$\pi_{\text{rdm}}$};
            \node[box, above right=9mm and -25mm of diff, fill=green!20] (demo) {Human demos\\$\mathcal{D}_{\text{demo}}$};
            \node[box, below=17mm of dec, fill=white] (unroll) {Unroll trajectory $\tau$};
            \node[box, below=of unroll, fill=purple!20] (data) {$\mathcal{D}\leftarrow\mathcal{D}\cup\{\tau\}$};

            \draw[->, thick] (start) -- (dec);
            \draw[->, thick] (demo) -- node[lbl, left] {train} (diff);
            \draw[->, thick] (dec) -- node[lbl, above left=-1mm] {\rrev{$u<\xi$}} (diff);
            \draw[->, thick] (dec) -- node[lbl, above right=-1mm] {\rrev{$u\geq \xi$}} (rdm);
            \draw[->, thick] (diff) |- (unroll);
            \draw[->, thick] (rdm)  |- (unroll);
            \draw[->, thick] (unroll) -- (data);
            \draw[->, thick] (data.east) -- ++(26mm,0) |- (dec.east);
        \end{tikzpicture}}
        \caption{\rev{Data Collection Procedure}}
        \label{fig:data-collection}
    \end{subfigure}

    \caption{LaGarNet: \textbf{(a)} the probabilistic graphical model and \textbf{(b)} the network architecture, where matching colours, line styles and nodes denote identical processes. $\bm{g}$ is the goal observation, $\bm{x}_t$ and $\bm{z}_t$ the observation and latent state before action $\bm{a}_t$, and $\bm{h}_t$ the hidden state of the recurrent backbone. In (a), purple observation and grey action nodes are model inputs and yellow action nodes are sampled during planning. In (b), the \textit{Prior} (red, solid) predicts $\boldsymbol{\tilde{z}}$ from $\boldsymbol{h}$ and $\boldsymbol{g}$, the \textit{Posterior} (orange, dashed) additionally integrates $\boldsymbol{x}$, and a joint decoder (blue) emits $\boldsymbol{\hat{y}}$ and $\boldsymbol{\hat{r}}$. \textbf{(c)} Sensitivity of the coverage-alignment reward $\mathcal{R}_{CA}$ to $\alpha_r$. \textbf{(d)} Data collection: a Diffusion Policy expert is mixed at ratio $\xi$ with the mask-biased random policy.}
    \label{fig:method-overview}
\end{figure*}

\rev{Our coverage-alignment reward $\mathcal{R}_{CA}$ combines the step-wise normalised coverage difference $\mathcal{R}_{dC}$ with the step-wise difference in maximum IoU between the current and goal cloth masks, $\irev{\mathcal{R}_{dU}}$:} 

\begin{equation}
    \mathcal{R}_{CA} = \alpha_r \cdot \mathcal{R}_{dC} + (1-\alpha_r) \cdot \irev{\mathcal{R}_{dU}},
\end{equation}
\rev{where \rrev{$\alpha_r = 0.3$ (Figure \ref{fig:reward-function}),} $\mathcal{R}_{dC} = \text{NC}_t - \text{NC}_{t-1}$ and $\irev{\mathcal{R}_{dU}} = \text{MaxIoU}_t - \text{MaxIoU}_{t-1}$; both $\mathcal{R}_{dC}$ and $\irev{\mathcal{R}_{dU}} \in [-1, 1]$, so that $\mathcal{R}_{CA} \in [-1, 1]$.} \resp{Being continuous and linear, it is easy to learn.}

\subsection{Proximal Expert and Random Policy Data Collection} \label{sec:data-collection}

\resp{Previous RSSM-based flatteners \cite{kadi2024planet} rely on a complex blend of data-collection policies.} \rev{\resp{Engineering flattening oracles for complex garments is extremely challenging, which hinders offline RL by preventing easy collection of trajectories around the successful region of the distribution.} LaGarNet mitigates these biases with a proximal expert policy and a mask-biased random policy (Figure \ref{fig:data-collection}).} \rev{In particular, we train a Diffusion Policy \cite{chi2023diffusionpolicy} on \resp{50} human demonstrations to replace the rigid oracle experts of previous methods \cite{kadi2024planet}; as Figure~\ref{fig:longsleeve-flattening} shows, \resp{it induces numerous nearly successful trajectories, diversifying the data around the near-successful region}.} \rev{For exploration, we introduce a mask-biased random policy rather than the fully random or corner-biased policies of prior work; it samples the pick pixel uniformly from the object mask and the place pixel from the image. Pairing the Diffusion Policy with this exploration at a 1:1 ratio curates a rich, targeted dataset.}

\section{Experiments} \label{sec:experiments}

\ral{Our experiments investigate three research questions: (1) how LaGarNet compares against state-of-the-art (SoTA) single-picker pick-and-place flattening baselines (Section \ref{sec:comp-sota}); (2) how important each component of LaGarNet is (Section \ref{sec:abl}); and (3) how LaGarNet performs in the real world (Section \ref{sec:real-exp}).}\ifwithappendix{} \ral{Appendix \ref{sec:comp-diff} additionally asks what advantage GC-RSSM brings over a Diffusion Policy.}\fi{} We evaluate our garment flattening policy with four standard metrics. \textbf{Normalised Coverage (NC)} is the ratio of the garment's current 2D projected area to its maximally flattened area, isolating the extent of unfolding and bounded to $[0,1]$. \resp{\textbf{Normalised Improvement (NI)} is the relative coverage increase from the initial configuration, $\frac{c_t - c_0}{c_{\text{max}} - c_0}$.} \textbf{Maximum Intersection-over-Union (Max IoU)} is the highest IoU between the current garment mask and the target flattened mask over rotation and translation. \textbf{Success Rate (SR)} is the percentage of trials reaching a state with both $\text{NC} > 90\%$ and $\text{Max IoU} > 80\%$.

\rev{\resp{We report both last-step and cumulative values over the horizon, since our question is whether the world model plans effectively, not whether the policy holds a flattened state.} Simulated episodes run 30 steps, showing how each policy behaves once the easy gains are exhausted. Real-world trials stop at 20 steps to reduce evaluation time (Section \ref{sec:real-exp})---a human operator comfortably flattens every garment type well within 20 actions.} In both settings each garment's goal state (its maximally flattened configuration) is captured beforehand, serving as the ground-truth reference and as input to goal-conditioned policies: in simulation we use the states of Canberk et al.~\cite{canberk2023cloth}, whilst in the real world we first flatten the garment \rrev{to a canonical state} and then manually crumple it\rev{\rrev{---every related test trial uses the same goal image but different initial states.}} \resp{Augmenting the goal and current images independently absorbs the slight garment--camera offset.}

\rev{We train all models for 50,000 update steps and validate every 1,000 steps to save the best checkpoint. We augment all policy inputs with appearance-level Gaussian noise, $SE(2)$ transformations of the current observation strictly coupled to the pixel action, and independent affine transformations of the goal observation. Simulation uses 200 training, 10 validation and 30 evaluation trials. \resp{We score checkpoints by the sum of average final NC and Max IoU, breaking ties above \irev{1.9 out of 2.0} in favour of fewer steps.}}

\subsection{Comparison of \text{LaGarNet} against Other Neural Controllers} \label{sec:comp-sota}

\begin{figure*}[t!]
\centering
\subfloat[\rev{Longsleeved T-shirt flattening against SoTA baselines.}]{%
    \includegraphics[width=0.6\linewidth]{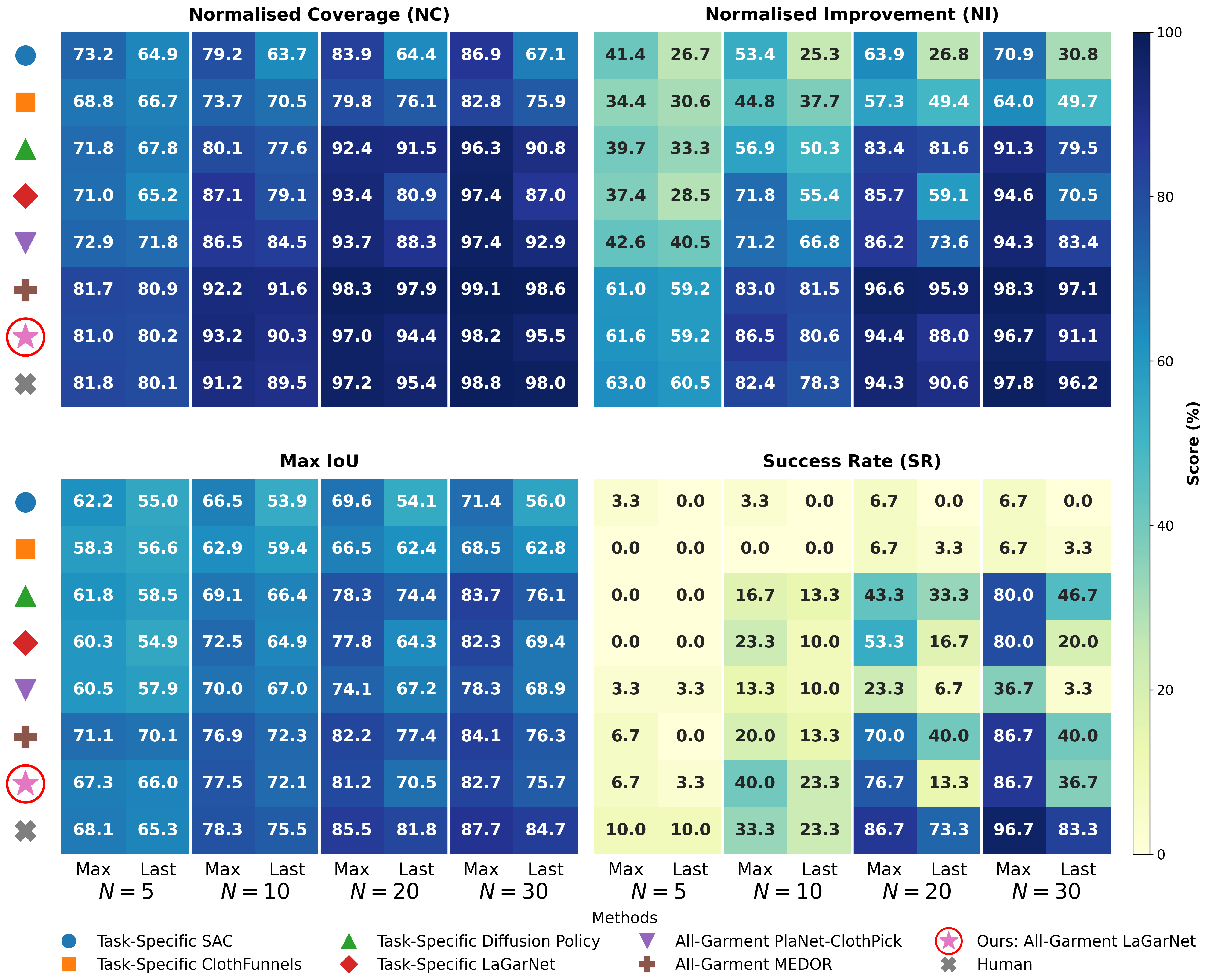}%
    \label{fig:longsleeve-flattening}}\hfill
\subfloat[\rev{All-garment generalisation by action step 20.}]{%
    \includegraphics[width=0.4\linewidth]{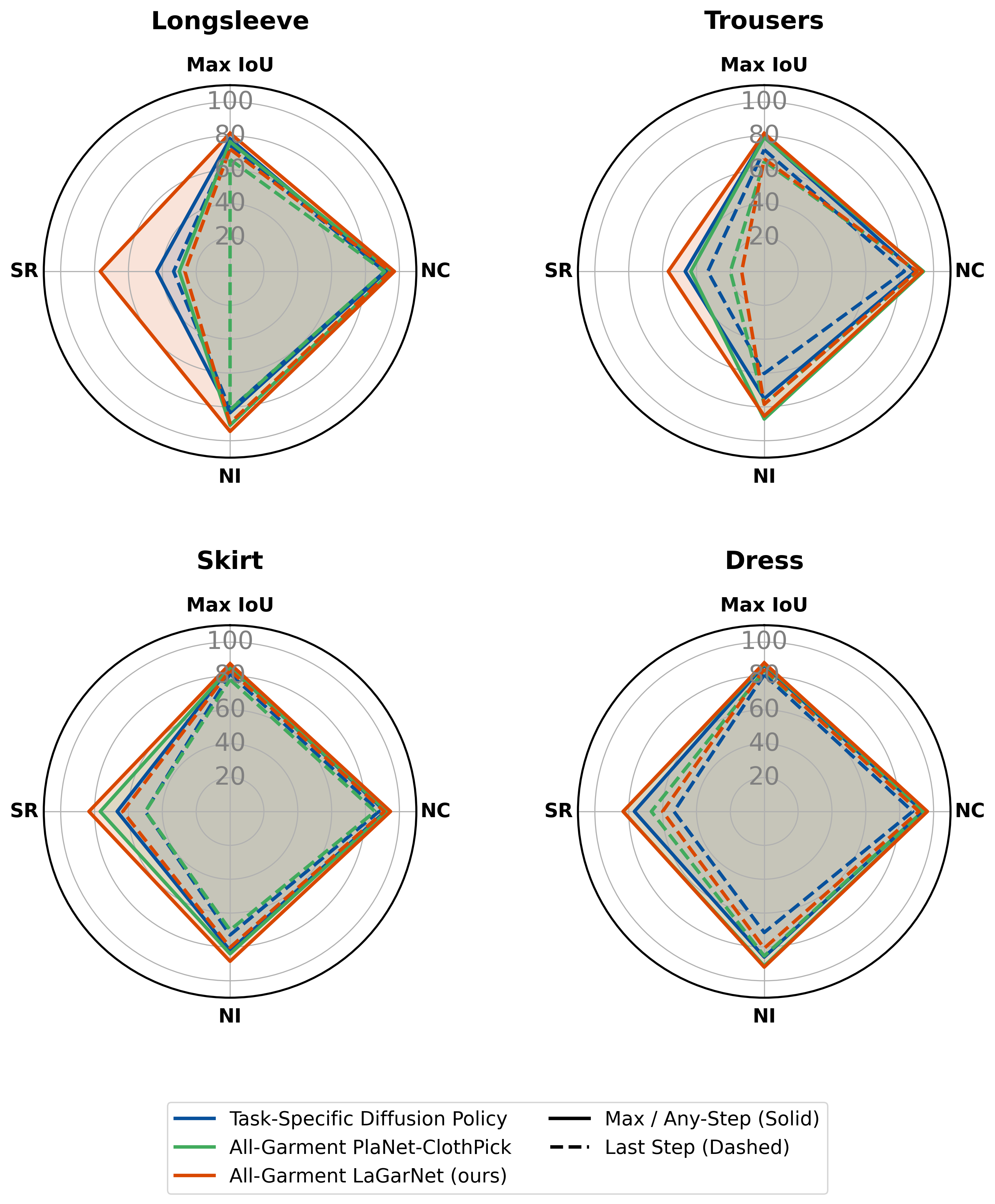}%
    \label{fig:sim-lagarnet-results}}
\caption{\ral{Simulation results for single-arm pick-and-place controllers. \textbf{(a)} Per-horizon longsleeve flattening; rows are methods and each horizon $N \in \{5, 10, 20, 30\}$ spans a \emph{Max} column (best within $N$ steps) and a \emph{Last} column (value at step $N$). LaGarNet matches the mesh-based planner MEDOR; its NC and Max IoU barely change between Max and Last, but SR drops sharply (Section \ref{sec:discussion}). \textbf{(b)} All four garment types by action step 20.}}
\label{fig:sim-results}
\end{figure*}

\begin{table}[t!]
    \centering
    \caption{\ral{Computational requirements and data efficiency across garment flattening baselines. Data is episodes $\times$ pick-and-places; TS is task-specific, AG all-garment; Par.\ is network parameters, Inf.\ inference time per action, Coll.\ data-collection time and Train training time. $\dagger$ marks values reported in the original papers; \resp{the all-garment 111\,h is $4 \times 9 + \max(75, 34)$}. Our measurements use an NVIDIA RTX 3090 Ti and an AMD Ryzen 9 5900X.}}
    \label{tab:efficiency-comparison}
    \footnotesize
    \setlength{\tabcolsep}{6pt}%
    \begin{tabular}{lccccc}
    \toprule
    \textbf{Method} & \textbf{Data} & \textbf{Par.} & \textbf{Inf.} & \textbf{Coll.} & \textbf{Train} \\
     & & \textbf{(M)} & \textbf{(s)} & \textbf{(h)} & \textbf{(h)} \\
    \midrule
    ClothFunnels    & 12.5k$\times$10$^\dagger$ & 12.6 & 2     & 48$^\dagger$ & 48$^\dagger$ \\
    MEDOR           & 5k$\times$100$^\dagger$   & 37.5 & 103   & 0.64         & 144$^\dagger$ \\
    Diffusion Policy (TS)         & 200$\times$20             & 83.7 & 1.5   & 9            & 35 \\
    PlaNet-ClothPick (AG)         & 5k$\times$20              & 6.3  & 4     & 111          & 31 \\
    LaGarNet (AG)   & 5k$\times$20              & 7.11 & 6     & 111          & 33 \\
    \bottomrule
    \end{tabular}
\end{table}

\subsubsection{Simulation Setup} \label{sec:simulation}

We develop \text{LaGarNet} in Canberk et al.'s simulation environment~\cite{canberk2023cloth} (SoftGym \cite{lin2020softgym}, PyFlex \cite{li2018learning}). \rev{Following the \text{DRAPER} strategy \cite{kadi2025draper}, we extend it to include parallel gripper misgrasping and multi-layer grasping. We evaluate simulated longsleeved T-shirts, trousers, dresses, and skirts from \text{Cloth3D} \cite{bertiche2020cloth3d} (Figure \ref{fig:sim-train-init}), scaled by 0.8. Rendered via a top-down camera (2\,m height, 45$^\circ$ field of view), each 1\,kg garment maintains Canberk et al.'s default physical parameters (stiffness, friction, damping). We evaluate on 30 ``hard'' (crumpled) initial states within the workspace.}

\subsubsection{Baselines} \label{sec:baseline} 

We benchmark our model-based RL method \textbf{LaGarNet} against RSSM-based \textbf{PlaNet-ClothPick} \cite{kadi2024planet} and the model-free \textbf{ClothFunnels} \cite{canberk2023cloth}. \rev{We also include the state-of-the-art mesh planner \textbf{MEDOR} \cite{huang2022mesh}, cropping observations to match its tuned garment-to-image ratio. Further baselines include the deep behaviour cloning method \textbf{Diffusion Policy} \cite{chi2023diffusionpolicy, kadi2025draper}, \textbf{SAC} \cite{haarnoja2018soft}, and a \textbf{Human} teleoperation baseline. Except for \text{ClothFunnels} and \text{MEDOR}, all trained methods use the \text{DRAPER} sim-to-real vision strategy \cite{kadi2025draper}. \resp{The \text{Diffusion Policy} baseline trains on 200 human demonstrations, four times what our data collector uses (Section \ref{sec:data-collection}), bounding what more demonstrations can contribute.} \text{LaGarNet} and \text{PlaNet-ClothPick} (which uses a different reward and no goal observations) learn from 5,000 generated episodes (100,000 transitional steps). All-garment variants use all four clothing types at task-specific data volumes.}

\subsubsection{Results} \label{sec:sota-result}

\rev{Figure \ref{fig:longsleeve-flattening} summarises flattening performance. All-Garment LaGarNet performs comparably to MEDOR and competitively with humans at intermediate horizons, though humans achieve the highest success rate at step 30 \irev{by more precisely aligning garments}. All-Garment LaGarNet outperforms the Task-Specific variant, demonstrating the architecture's benefit from morphological diversity \irev{(Figure \ref{fig:sim-lagarnet-results}; Section \ref{sec:discussion})}. Given 30 steps, Diffusion Policy recovers to competitive results despite weak short-horizon behaviour. Figure \ref{fig:sim-lagarnet-results} shows \text{LaGarNet} exceeding 95\% NC across all garment types at step 20, with higher success on simpler topologies indicating a slight bias. \resp{Table \ref{tab:efficiency-comparison} shows the efficiency trade-off: MEDOR reaches high Max IoU at $\sim$103\,s per step and 37.5\,M parameters, and \text{Diffusion Policy} is fast but needs 83.7\,M with lower success rates; All-Garment \text{LaGarNet} needs 7.11\,M and $\sim$6\,s, facilitating real-world deployment.}}

\subsection{Ablation Study on \text{LaGarNet}} \label{sec:abl}

\begin{figure*}[t!]
    \centering

    \subfloat[\centering Latent Dynamic and Goal Ablation \label{fig:goal-ablation}]{
        \includegraphics[width=0.215\linewidth]{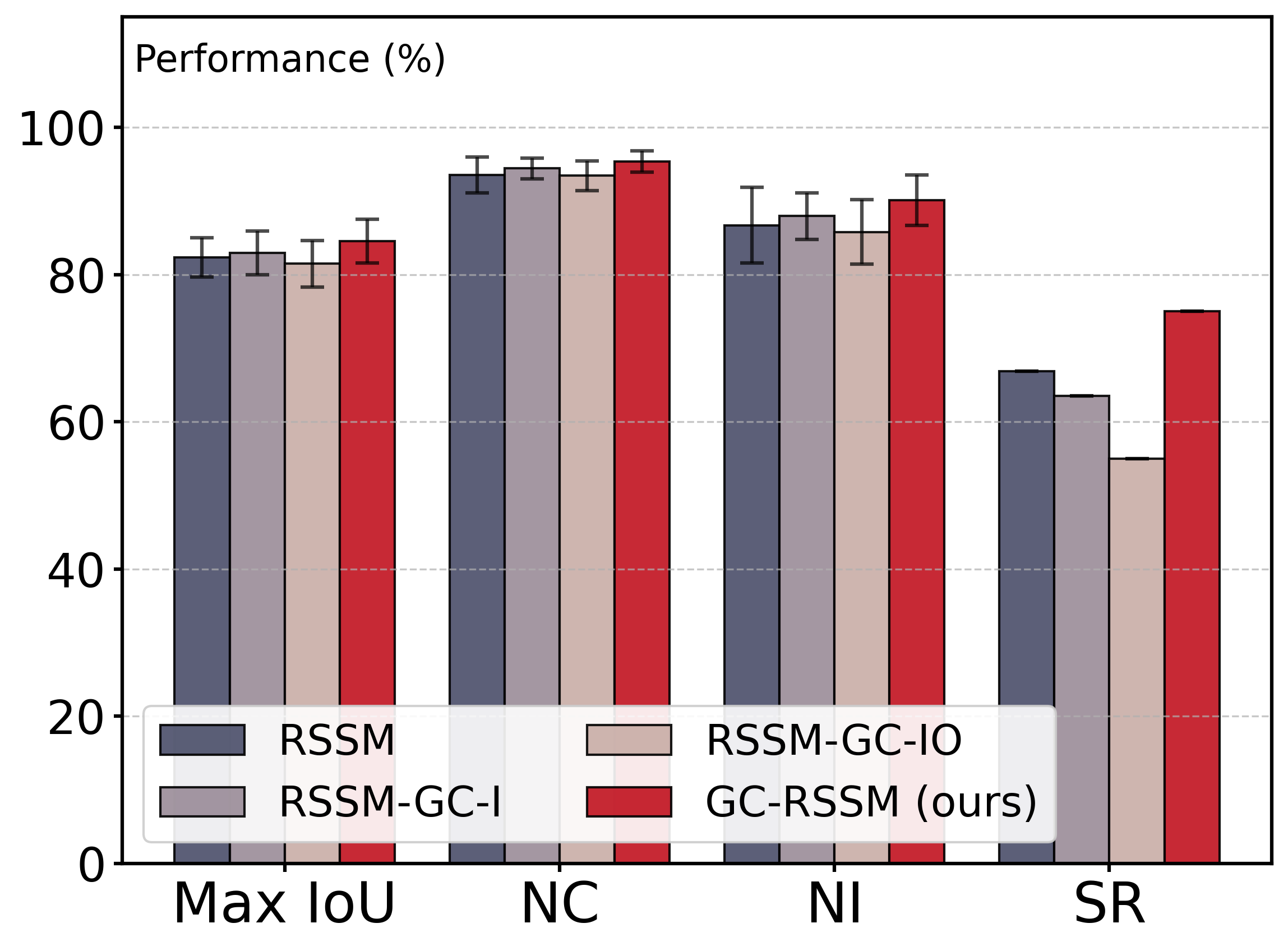}
    }\hfill
    \subfloat[\centering Reward Ablation \label{fig:reward-ablation}]{
        \includegraphics[width=0.215\linewidth]{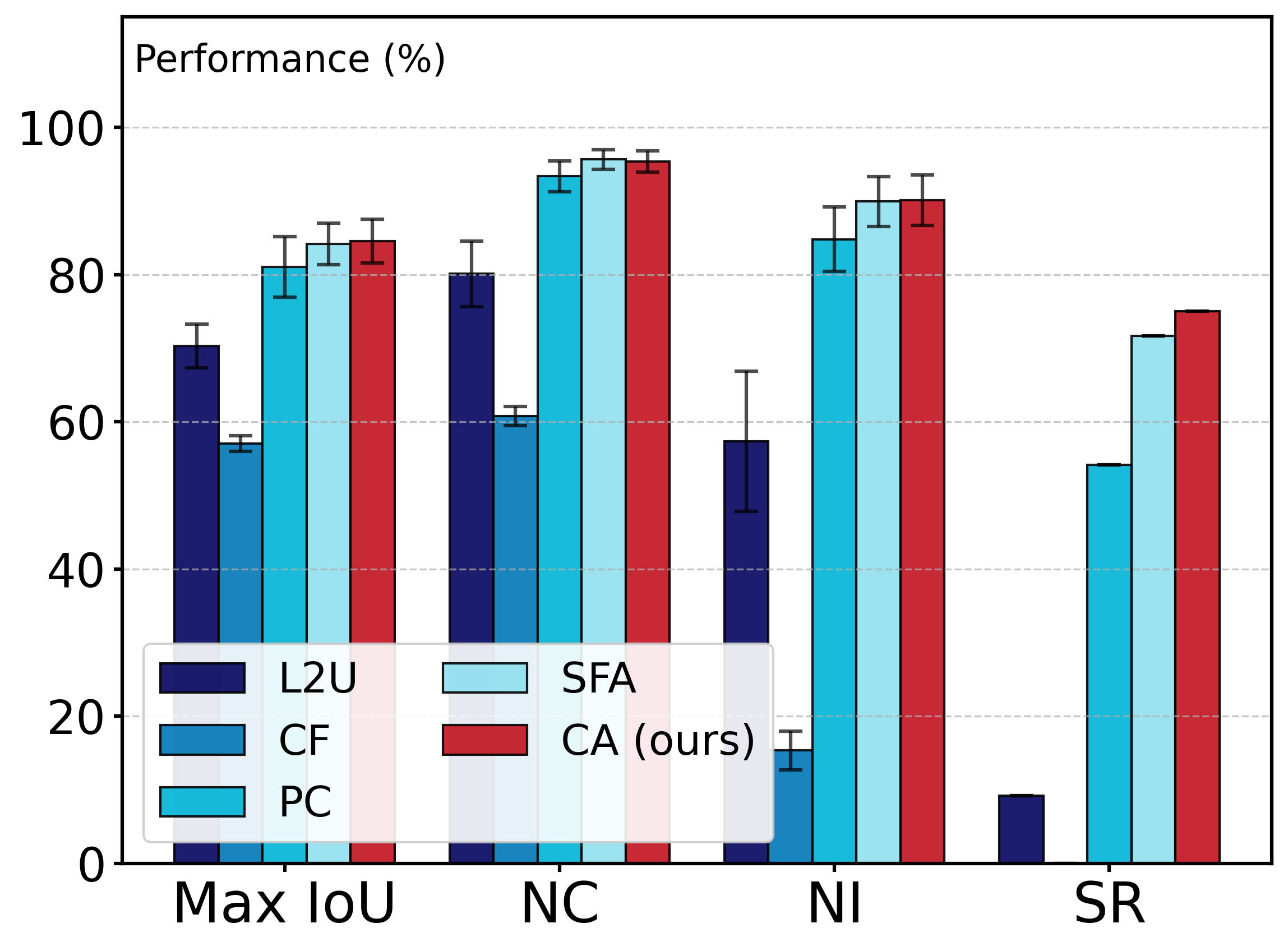}
    }\hfill
    \subfloat[\centering \rev{Data Size Ablation} \label{fig:datasize-ablation}]{
        \includegraphics[width=0.215\linewidth]{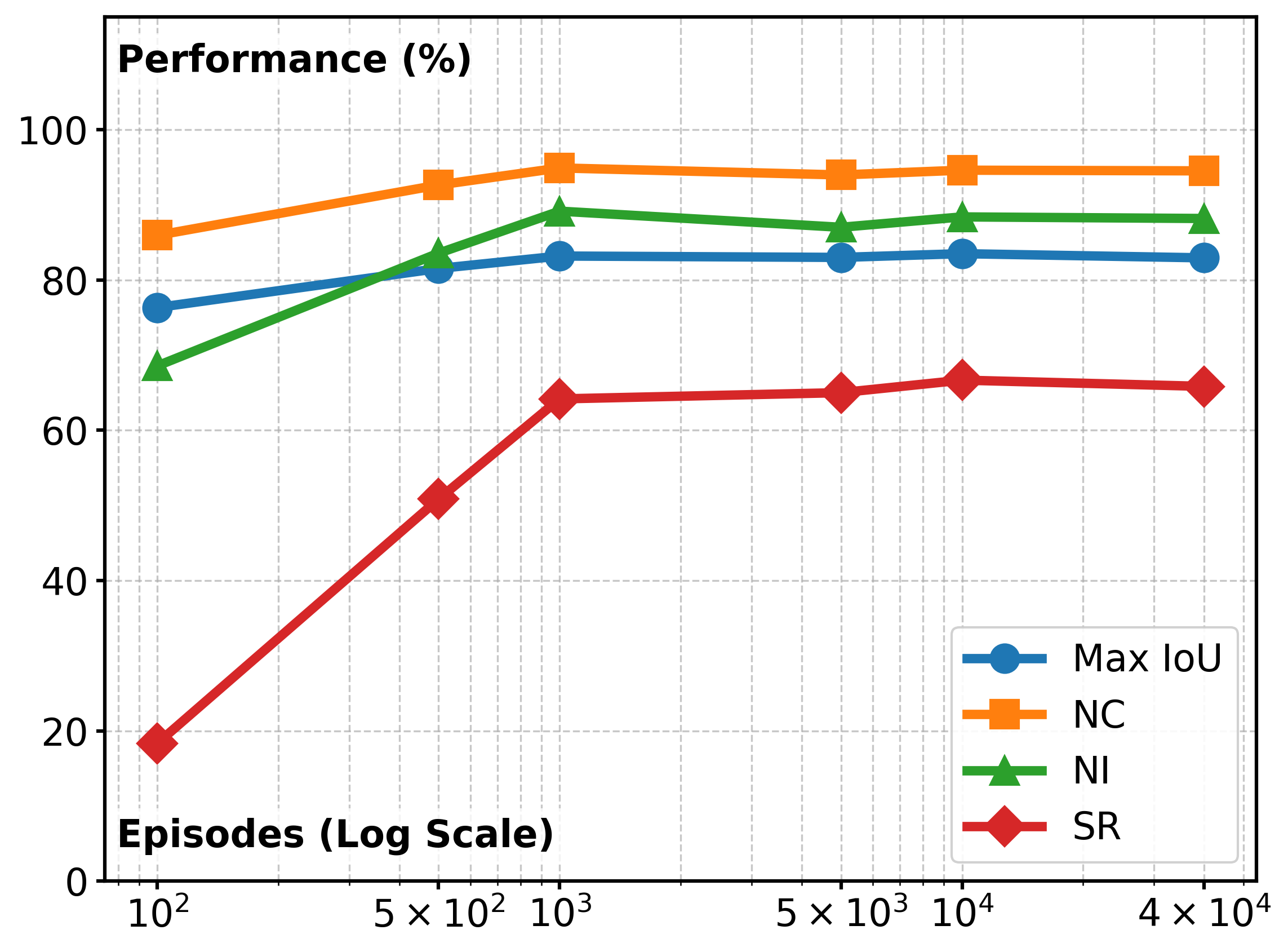}
    }\hfill
    \subfloat[\centering Data Composition Ablation \label{fig:dataratio-ablation}]{
        \includegraphics[width=0.215\linewidth]{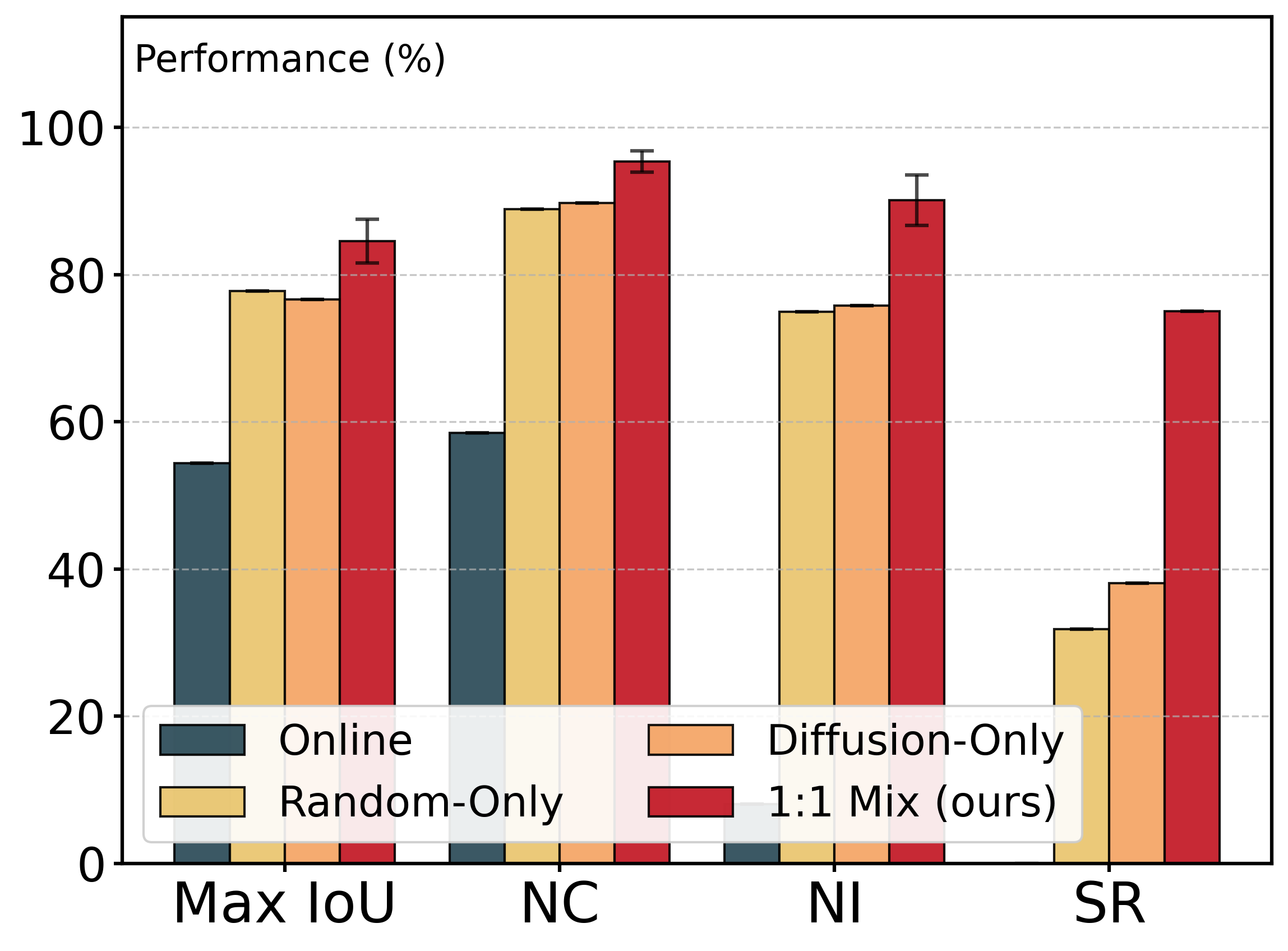}
    }

    \vspace{0.15cm}

    \begin{minipage}[b]{0.96\textwidth}
        \subfloat[\centering \rev{MPC Planning Ablation} \label{fig:horizon-ablation}]{%
            \begin{minipage}[b]{0.47\textwidth}
            \centering
            \includegraphics[width=\linewidth]{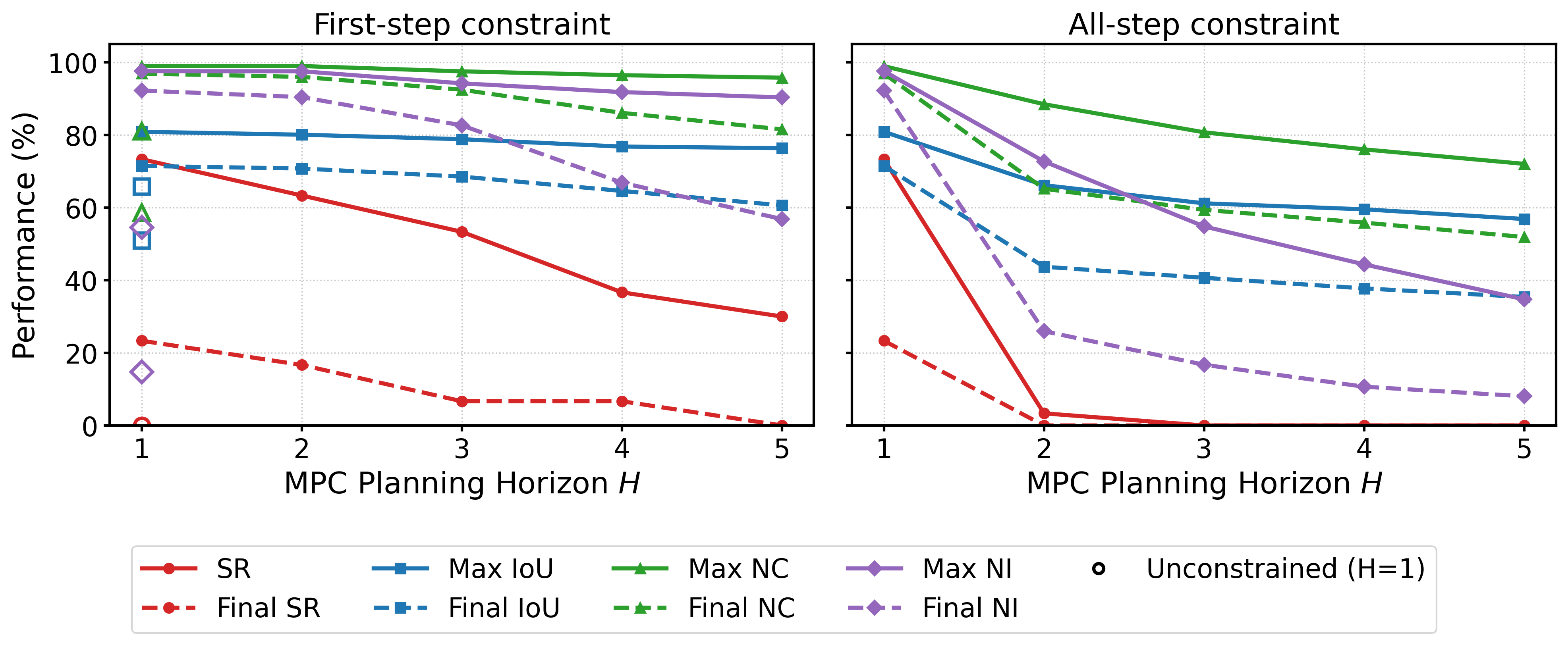}
            \end{minipage}%
        }\hfill
        \subfloat[\centering \rev{Reconstruction quality (MSE, SSIM)} \label{tab:recon-quality}]{%
            \begin{minipage}[b]{0.46\textwidth}
            \centering

            \centering
            \resizebox{\linewidth}{!}{%
            \setlength{\tabcolsep}{2pt}%
            \renewcommand{\arraystretch}{1.1}%
            \begin{tabular}{l|cc|cc|cc|cc}
             & \multicolumn{2}{c|}{Posterior} & \multicolumn{2}{c|}{Prior 1} & \multicolumn{2}{c|}{Prior 2} & \multicolumn{2}{c}{Prior 3} \\
            \hline
             Model, Metrics & MSE & SSIM   & MSE & SSIM   & MSE & SSIM   & MSE & SSIM  \\
            \hline
            RSSM & \textbf{0.26} & \textbf{94.09} &1.92 & 63.37 &2.60 & 62.00 &2.91 & 52.17\\
            RSSM, GC-I & 0.28 & 94.72 &1.87 & 65.58 &2.59 & 58.46 &2.80 & 56.05\\
            RSSM, GC-IO & 0.37 & 92.78 & \textbf{1.57} & \textbf{68.12} & \textbf{2.34}  & 62.10 &2.70 & 54.17\\
            GC-RSSM & 0.32 & 94.03 &1.67 & 67.72 &2.40 & \textbf{63.27} & \textbf{2.70} & \textbf{57.66}\\
            \bottomrule
            \end{tabular}%
            }
            \end{minipage}%
        }
    \end{minipage}

    \caption{\ral{Ablation study for LaGarNet in simulation. \textbf{(a--d)} Component ablations on the all-garment models. \textbf{(e)} Flattening performance against the MPC planning horizon $H$, cumulative (\emph{Max}) and final-step. Left: the default planner, constraining only the first action to the observed cloth mask; hollow markers are the unconstrained-CEM variant at $H=1$. Right: the all-step variant, additionally projecting actions $k \geq 1$ onto the predicted mask for step $k$. \textbf{(f)} Reconstruction quality of the models of (a) by MSE and SSIM, both $\times$100 over 10 trials per garment; GC-RSSM blurs the multi-step prior reconstructions least.}}
    \label{fig:ablation-and-recon-combined}
\end{figure*}

\rev{We ablate LaGarNet's latent dynamics, reward, data size, dataset mixture, and planning horizon using Section \ref{sec:simulation}'s setup. Experiments use RGB-to-Mask variants, except latent dynamics ablations (RGB-to-RGB) to assess prediction quality. Models are evaluated on 30 held-out hard states across four arenas. The planning ablation isolates the hardest longsleeve arena, \irev{evaluates it over 30 steps}, and logs planning time.}

\subsubsection{Baselines}

\rev{Latent dynamics ablations compare \textbf{GC-RSSM} and \textbf{RSSM} (Section \ref{sec:gc-rssm}) against \textbf{RSSM-GC-I} (decodes current image only) and \textbf{RSSM-GC-IO} (predicts both current and goal images), separating goal-conditioning from GC-RSSM's structure. Reward ablations test \textbf{L2U} \cite{gu2023learning}, \textbf{CF} \cite{canberk2023cloth}, \textbf{PC} \cite{kadi2024planet}, and \textbf{SFA} \cite{avigal2022speedfolding} (Section \ref{sec:related-work}). Dataset ablations vary volume, composition, and online exploration noise \cite{hafner2019learning}. The planning ablation sweeps \irev{$H \in \{1,2,3,4,5\}$} and tests an \textbf{unconstrained} variant ($H=1$, no mask rejection). \rrev{We also add \textbf{all-step} variants constraining actions to predicted masks at step $k$, projecting empty populations onto nearest mask pixels.}}

\subsubsection{Results}

Figure \ref{fig:goal-ablation} shows gains stem from GC-RSSM's structure, not just goal-conditioning. RSSM-GC-I performs similarly, but RSSM-GC-IO performs poorly due to increased learning burden. GC-RSSM yields a better prior latent representation for planning (Figure \ref{tab:recon-quality}).\ifwithappendix{} \ral{Appendix \ref{sec:recon-appendix} shows the reconstructions themselves.}\fi{} \resp{Figure \ref{fig:reward-ablation} shows the reward is decisive: success rate falls to 53.9\%, 9.0\% and 0.0\% under the PlaNet-ClothPick, Learning2Unfold and ClothFunnels rewards, and to 71.4\% under SpeedFolding's.} Performance plateaus beyond 1,000 episodes (Figures \ref{fig:datasize-ablation}, \ref{fig:dataratio-ablation}), confirming the importance of the data-collection strategy. The greedy one-step planner dominates (Figure \ref{fig:horizon-ablation}): success falls monotonically from 73.3\% ($H=1$) to 30.0\% ($H=5$). Lookahead is therefore detrimental here. \irev{Longer horizons also increase mean planning time from 4.17\,s ($H=1$) to 5.33\,s ($H=5$).} Dropping mask rejection at $H=1$ collapses success to near zero as samples miss the garment. \rrev{Extending constraints to every step deepens this loss (Figure \ref{fig:horizon-ablation}, right). Ablations rank mask-constrained action sampling as the primary performance driver, followed by the reward, and finally the architecture.}

\subsection{Transfer Ability of LaGarNet to Real-World Settings} \label{sec:real-exp}

\begin{table*}[t!]
\caption{\ral{Real-world garment flattening performance.} Each block is an evaluation horizon and every value is cumulative over it: NC, NI and Max IoU give the best reached within the first 10 or 20 actions, and SR the fraction of trials meeting both criteria at any step. \ral{LaGarNet improves substantially on PlaNet-ClothPick but still falls short of human performance.}}
\label{tab:real-worldflattening}
\centering
\resizebox{0.96\textwidth}{!}{%
\renewcommand{\arraystretch}{1.1}%
\setlength{\tabcolsep}{2pt}%
\begin{tabular}{c|cccc|cccc||cccc}
Method & \multicolumn{4}{c|}{All-Garment PlaNet-ClothPick \cite{kadi2024planet}}  & \multicolumn{4}{c||}{All-Garment LaGarNet} & \multicolumn{4}{c}{Human Policy} \\
\cline{1-13}
10 Steps & NC $\uparrow$  & NI $\uparrow$ & Max IoU $\uparrow$    & SR  $\uparrow$ 
& NC $\uparrow$  & NI $\uparrow$ & Max IoU $\uparrow$     & SR  $\uparrow$ 
& NC $\uparrow$  & NI $\uparrow$ & Max IoU $\uparrow$    & SR  $\uparrow$ \\
\hline
Longsleeve & 67.1 $\pm$ 12.7 & 30.4 $\pm$ 20.9 & 63.2 $\pm$ 10.3 & 0/10 & 73.4 $\pm$ 8.2 & 52.0 $\pm$ 16.2 & 66.3 $\pm$ 7.0 & 1/10 & 89.1 $\pm$ 7.8 & 81.0 $\pm$ 13.8 & 78.5 $\pm$ 6.3 & 4/10 \\
Trousers & 60.6 $\pm$ 16.4 & 24.1 $\pm$ 21.9 & 57.9 $\pm$ 14.9 & 1/10 & 78.7 $\pm$ 10.8 & 57.2 $\pm$ 17.0 & 70.1 $\pm$ 9.2 & 3/10 & 96.4 $\pm$ 5.5 & 93.0 $\pm$ 10.9 & 85.7 $\pm$ 7.7 & 9/10 \\
Dress & 71.9 $\pm$ 19.5 & 74.6 $\pm$ 28.8 & 69.5 $\pm$ 16.4 & 3/10 & 99.0 $\pm$ 2.2 & 97.6 $\pm$ 6.0 & 88.6 $\pm$ 2.7 & 10/10 & 94.7 $\pm$ 12.0 & 91.3 $\pm$ 19.8 & 85.3 $\pm$ 8.5 & 9/10 \\
Skirt & 76.1 $\pm$ 17.1 & 44.0 $\pm$ 26.1 & 74.1 $\pm$ 15.5 & 4/10 & 87.7 $\pm$ 11.8 & 75.7 $\pm$ 22.2 & 83.9 $\pm$ 9.5 & 8/10 & 95.4 $\pm$ 6.8 & 85.9 $\pm$ 24.7 & 91.0 $\pm$ 5.1 & 8/10 \\
\hline
Real Total & 69.0 $\pm$ 17.6 & 43.3 $\pm$ 31.4 & 66.2 $\pm$ 15.7 & 8/40 & 84.7 $\pm$ 13.3 & 70.6 $\pm$ 24.3 & 77.2 $\pm$ 12.0 & 22/40 & 93.9 $\pm$ 8.8 & 87.8 $\pm$ 18.7 & 85.1 $\pm$ 8.3 & 30/40 \\
\hline
\hline
20 Steps 
& NC $\uparrow$  & NI $\uparrow$ & Max IoU $\uparrow$    & SR  $\uparrow$ 
& NC $\uparrow$  & NI $\uparrow$ & Max IoU $\uparrow$     & SR  $\uparrow$ 
& NC $\uparrow$  & NI $\uparrow$ & Max IoU $\uparrow$    & SR  $\uparrow$ \\
\hline
Longsleeve & 69.2 $\pm$ 13.2 & 34.6 $\pm$ 22.1 & 64.5 $\pm$ 10.3 & 0/10 & 84.5 $\pm$ 7.9 & 71.1 $\pm$ 17.0 & 74.1 $\pm$ 6.8 & 4/10 & 96.8 $\pm$ 7.6 & 94.2 $\pm$ 13.9 & 87.6 $\pm$ 5.8 & 9/10 \\
Trousers & 60.6 $\pm$ 16.4 & 24.1 $\pm$ 21.9 & 57.9 $\pm$ 14.9 & 1/10 & 80.2 $\pm$ 10.3 & 60.1 $\pm$ 16.9 & 70.5 $\pm$ 9.0 & 3/10 & 100.0 $\pm$ 0.0 & 100.0 $\pm$ 0.0 & 93.6 $\pm$ 0.7 & 10/10 \\
Dress & 75.8 $\pm$ 19.5 & 90.6 $\pm$ 17.7 & 72.5 $\pm$ 16.2 & 3/10 & 100.0 $\pm$ 0.0 & 100.0 $\pm$ 0.0 & 89.5 $\pm$ 1.9 & 10/10 & 100.0 $\pm$ 0.0 & 100.0 $\pm$ 0.0 & 89.2 $\pm$ 1.7 & 10/10 \\
Skirt & 80.4 $\pm$ 17.1 & 54.0 $\pm$ 31.7 & 77.4 $\pm$ 14.7 & 5/10 & 96.5 $\pm$ 3.2 & 91.4 $\pm$ 9.8 & 90.4 $\pm$ 2.3 & 10/10 & 97.4 $\pm$ 5.7 & 90.3 $\pm$ 24.2 & 92.6 $\pm$ 4.2 & 9/10 \\
\hline
Real Total & 71.5 $\pm$ 18.3 & 50.8 $\pm$ 34.9 & 68.1 $\pm$ 16.0 & 9/40 & 90.3 $\pm$ 10.6 & 80.7 $\pm$ 20.4 & 81.1 $\pm$ 10.7 & 27/40 & 98.6 $\pm$ 5.0 & 96.1 $\pm$ 14.5 & 90.7 $\pm$ 4.4 & 38/40 \\
\bottomrule
\end{tabular}%
}
\end{table*}

\begin{table}[t!]
    \centering
    \caption{\ral{Trajectory and step-wise failure rates for LaGarNet in the real world; a trajectory succeeds if it reaches the success state within 20 steps. ``Messing-near-success state'' counts drops in the step right after a near-success step, and ``As-if-manipulating-other-object'' counts steps treating part of the garment as another topology, most often a trouser leg as a sleeve.}}
    \label{tab:failure-rates}
    \footnotesize
    \begin{tabular*}{\columnwidth}{@{\extracolsep{\fill}}llc@{}}
    \toprule
    \textbf{Failure Category} & \textbf{Specific Mode} & \textbf{Rate} \\ \midrule
    \multirow{4}{*}{\textbf{Trajectory}}
    & Fold Underneath & 3/40 \\
    & Fold On the Top & 2/40 \\
    & Cannot untwist & 5/40 \\
    & Cannot stabilise on success state & 4/40 \\ \midrule
    \multirow{4}{*}{\textbf{Step-wise}}
    & Segmentation failure & 1/800 \\
    & Misgrasping failure & 3/800 \\
    & Messing-near-success state & 168/298 \\
    & As-if-manipulating-other-object & 28/800 \\ \bottomrule
    \end{tabular*}
\end{table}

\rev{Our zero-shot real-world study evaluates simulation-trained LaGarNet and PlaNet-ClothPick (Section \ref{sec:sota-result}) across four garment types (Figure \ref{fig:real-garments}). \resp{We exclude MEDOR for its prohibitive execution time and Diffusion Policy for its workspace-constrained inference.} \resp{We additionally evaluate a soft, high-friction longsleeved dress as an out-of-distribution (OOD) topology.}} \rev{\resp{We use a UR5e arm with an OnRobot RG2 parallel gripper (workspace radius 0.2--0.7\,m) and a RealSense D435i camera at $P_c = (0.03, -0.86, 1.66)$\,m, cropping its frames to a $450 \times 710$ region; everything runs on an RTX 3080 laptop.} \rrev{Because garments are visually flattened in practice, we relax real-world success criteria to NC $\geq$ 85 and Max IoU $\geq$ 75, \resp{absorbing camera distortion.}}}

\subsubsection{Results}
Table \ref{tab:real-worldflattening} shows the unified all-garment LaGarNet flattens longsleeved T-shirts, trousers, skirts and dresses in the real world, substantially improving on PlaNet-ClothPick. \ral{It trails the human baseline on narrow-appendage garments, yet stays competitive with the single-gripper mesh-based planners VCD \cite{lin2022learning} and MEDOR \cite{huang2022mesh}: \resp{within 20 steps it reaches} 84.5$\pm$7.9\% NC on longsleeved T-shirts against VCD's 77.3$\pm$14.1\%, and \resp{100.0$\pm$0.0}\% NI on dresses against MEDOR's 46.8\%. On an unseen longsleeved dress of soft, high-friction material it reaches only 77.2$\pm$6.5\% NC, 42.1$\pm$22.6\% NI and 66.4$\pm$3.2\% Max IoU, succeeding in 0 of 10 trials. Topology dependence is strong: a perfect 10/10 success rate on dresses and skirts within 20 steps, but only 3/10 and 4/10 on trousers and longsleeved shirts.}

\paragraph{Limitations and Failure Cases}
\rev{LaGarNet faces a persistent sim-to-real gap \cite{huang2022mesh, lin2022learning} and multi-task capacity issues: the unified policy underfits complex topologies and overfits simpler ones (Table \ref{tab:real-worldflattening}). It struggles to untwist narrow sections, often manipulating trouser legs like T-shirt sleeves (Table \ref{tab:failure-rates}).\ifwithappendix{} \ral{Appendix \ref{sec:failure-appendix} shows these cases.}\fi{} Furthermore, inaccurate first-step prior reward predictions from GC-RSSM occasionally cause the policy to disrupt already flattened near-success states (Table \ref{tab:failure-rates}). \resp{Prioritising geometry over physical properties, LaGarNet struggles with high-friction, low-stiffness garments, and the single-arm gripper limits unfolding stubborn wrinkles.} LaGarNet also cannot distinguish front from back---a challenging task even for human \irev{\cite{klein1983predressing}}. \resp{It also assumes a one-off canonical flattened goal image per garment.} \resp{Finally, a single primitive step takes 24.5\,s (perception 8.19\,s, inference 7.05\,s, execution 9.20\,s), resulting in over 8 minutes per 20-step episode and limiting industrial deployment.}}

\section{Discussion} \label{sec:discussion}

To our knowledge LaGarNet is the first integrated latent world model system to flatten four garment topologies with a single policy in a quasi-static setup\resp{, a result we credit to mask-constrained sampling, the coverage-alignment reward and the data-collection strategy as much as to the architecture (Section \ref{sec:abl})}. Applying an RSSM here raises two challenges: cloth deformation is complex, and high-level primitives make inferred states far less continuous than in velocity or position control. We find that RSSM \cite{hafner2019learning}, with our goal-conditioned extension, still learns the latent dynamics of cloth deformation in quasi-static setups. 

\ral{LaGarNet plans in a compact latent space instead of over a reconstructed mesh, making it more data-efficient and stable than model-free and online RL. It also predicts each candidate action's outcome, so the policy is explainable and does not depend on high-quality demonstrations.} Offline RL suffers from distribution shift \cite{prudencio2023survey}, and most methods train on data from a single policy \cite{agarwal2020optimistic}. \ral{We instead pair a mask-biased random policy with Diffusion Policy \cite{chi2023diffusionpolicy}, a BC method that learns to flatten from only 50 human demonstrations whilst collecting the near-success data a random policy rarely reaches.} \resp{We use BC only as a data collector; our deployed method never imitates an action, because BC requires extensive and expensive human expert data for garment manipulation.} BC also struggles to transfer to slightly different domains; \resp{our workspace constraints shift with the heuristic's crop, enforcing which during action denoising is difficult. LaGarNet simply rejects out-of-workspace samples while planning, so a loosely constrained simulation policy still transfers.}

\ral{LaGarNet's chief limitation is that it does not know when to stop.} It reaches a success state in 76.7\% of simulated longsleeve trials within 20 steps but holds it at the final step in only 13.3\% (Figure \ref{fig:longsleeve-flattening}), and in the real world 168 of 298 near-success steps end in a drop (Table \ref{tab:failure-rates}). The reward predictor is least reliable once the garment lies flat, so the planner continues acting when it should stop; human operators stop as soon as the garment looks flat, which accounts for much of their advantage at long horizons. A learned stopping criterion would close this gap. Nonetheless, this research opens an exciting direction for state-space models in deformable object manipulation, such as handling dough, clay, or cables. Although we focus on reward-based planning for flattening, the model also predicts future features, which could extend to goal-directed folding.

\section{Conclusion} \label{sec:conclusion}

We introduce LaGarNet, a general learning-based policy for single-arm pick-and-place (PnP) garment flattening. It successfully flattens four complex garment types (longsleeved T-shirts, trousers, skirts, and dresses) with a single policy in both simulated and real-world environments. We achieve this via a goal-conditioned recurrent state-space model, a targeted data-collection strategy, and a coverage-alignment reward. \rev{LaGarNet matches state-of-the-art mesh-based planners at a fraction of their parameter count and inference cost, marking the first successful application of state-space models to complex garments. We demonstrate that recurrent latent world models remain effective in the quasi-static PnP regime despite temporally discontinuous observations.}

\ifshowauthors
\section*{Acknowledgment}
\rev{We thank the Institute of Safe Autonomy at the University of York for hosting the experimental evaluations, and \textbf{Dr James Hilder} and \textbf{Dr Rob Woolley} for their help in setting up the robot platform; \textbf{Dr Stuart Norcross} at the University of St Andrews for access to essential GPU clusters; and \textbf{Dr Ivan Kapelyukh} at Imperial College London for insightful discussions on generalising this approach to dual-arm manipulation and cloth folding. We acknowledge the use of the Viking cluster at York, and of Google's Gemini 3 Flash Image to remove the background and enhance the image quality of the setup shown in Figure 1(c). This work was supported by the UK Engineering and Physical Sciences Research Council under grant EP/W014688/2 (DIGI-CORTEX).}
\fi

\ifwithappendix
\clearpage

\appendices

\section{Network Architecture, Planning and Hyperparameters} \label{sec:architecture}

\rev{This appendix specifies the exact architecture of LaGarNet, its test-time planning procedure, and all training hyperparameters; see Section \ref{sec:gc-rssm} for the conceptual overview.}

\begin{table*}[h]
    \centering
    \caption{Hyperparameters for LaGarNet and Diffusion Policies. GradClip represents gradient clipping. KL denotes Kullback-Leibler divergence, used for latent state regularisation. MPC stands for Model Predictive Control, utilised for trajectory planning. EMA refers to Exponential Moving Average, used to stabilise training weights. \label{tab:all-hyperparameters}}
   
    \begin{minipage}[t]{0.48\textwidth}
        
        \centering
        \textbf{(a) LaGarNet.} Other parameters are provided in Table (c). \\[0.15cm]
        \begin{tabular}{@{} 
            >{\raggedright\arraybackslash}p{0.50\linewidth} 
            >{\raggedright\arraybackslash}p{0.40\linewidth} @{}}
            \toprule
            \textbf{Hyperparameter} & \textbf{Value} \\
            \midrule
            $\bm{h}, \bm{z}$ Dimensions & 300, 60 \\
            Hidden Dimension & 300 \\
            $\beta_{KL}, \gamma_{KL}, \irev{\gamma_{KLo}}, \gamma_{ro}, \gamma_y, \gamma_r$ & 0.8, 1.0, 0.1, 0.1, 1.0, 1.0 \\
            $\sigma_{min}$, $c$ & 0.1, 1 \\
            $\bm{a}_{\text{no-op}}$ & $\bm{1}$ \\
            MPC Horizon / Candidates/ Iterations & 1 / 5000 / 100 \\
            Backbone Network & GRU RNN \cite{cho2014properties} \\
            \bottomrule
        \end{tabular}
        
        \vspace{0.6cm}
        
        \textbf{(b) Diffusion Policy for Data Collection.} Other parameters are the same as GC-DP in Table (c).\\[0.15cm]
        \begin{tabular}{@{} 
            >{\raggedright\arraybackslash}p{0.45\linewidth} 
            >{\raggedright\arraybackslash}p{0.45\linewidth} @{}}
            \toprule
            \textbf{Hyperparameter} & \textbf{Value} \\
            \midrule
            Denoising Iterations & 100 \\
            Noise Scheduler & iDDPM \cite{nichol2021improved} \\
            EMA Power & 0.75 \\
            Backbone Network & 1D Conditional U-Net \cite{chi2023diffusionpolicy, ronneberger2015u} with disabled Updown \cite{kadi2025draper} \\
            \bottomrule
        \end{tabular}
    \end{minipage}
    \hfill
    \begin{minipage}[t]{0.48\textwidth}
        \centering
        \textbf{(c) Comparison Configurations.}  \\[0.15cm]
        \begin{tabular}{@{} 
            >{\raggedright\arraybackslash}p{0.22\linewidth} 
            >{\raggedright\arraybackslash}p{0.21\linewidth} 
            >{\raggedright\arraybackslash}p{0.21\linewidth} 
            >{\raggedright\arraybackslash}p{0.21\linewidth} @{}}
            \toprule
            \textbf{Hyperparameter} & \textbf{LaGarNet} & \textbf{*-Comp} & \textbf{GC-DP} \\
            \midrule
            \textbf{$\bm{e}_x$} Dimension & 1024 & 512 & 512 \\
            \textbf{$T_h$} & 49 & 2 & 0 \\
            \textbf{$T_f$} & 1  & 2 & 1 \\
            Sequence Size & 49+1=50 & 2+2=4 & 0+1=1 \\
            Batch Size & 50 & 50 & 64 \\
            Input Resolution & $64 \times 64$ & $64 \times 64$ & $96 \times 96$ \\
            Vision Encoder & PlaNet's CNN \cite{hafner2019learning} & PlaNet's CNN & ResNet-18 \cite{he2016deep} with GroupNorm \cite{wu2018group} \\
            Optimisation & Adam ($\alpha$=$10^{-3}$, $\epsilon$=$10^{-4}$, $\lambda$=0) with GradClip (1000). & AdamW ($\alpha$=$10^{-4}$, $\lambda$=$10^{-6}$, $\epsilon$=$10^{-8}$), Cosine Warmup (500) with GradClip (100)  & AdamW ($\alpha$=$10^{-4}$, $\lambda$=$10^{-6}$, $\epsilon$=$10^{-8}$), Cosine Warmup (500) \\
            \bottomrule
        \end{tabular}
    \end{minipage}%
\end{table*}

\subsection{Image Encoder \& Decoder} The visual encoder processes the $64 \times 64$ RGB images (either the current observation $\bm{x}_t$ or the goal observation $\bm{g}$) into a 1024-dimensional continuous embedding, denoted as $\bm{e}_t$ and $\bm{e}_g$ respectively. The network comprises four sequential convolutional blocks. For the $64 \times 64$ resolution, the blocks have channel sizes of $\{32, 64, 128, 256\}$. Each block utilises a kernel size of $4 \times 4$ and a stride of $2$. The output of the final convolutional layer is flattened and projected via a linear layer to match the designated embedding dimension.

The observation decoder acts as a generative model $p(\hat{\bm{x}}_t \;|\; \bm{h}_t, \bm{z}_t)$ to reconstruct the mask image from the latent space, which is primarily used to compute the observation reconstruction loss. It takes the concatenated deterministic state $\bm{h}_t$ and stochastic state $\bm{z}_t$ (sampled from the posterior $\hat{\bm{z}}_t$ during training) as input, processes it through a linear dense layer, and reshapes it into a $1 \times 1$ spatial feature map. This is followed by four transposed convolutional layers with channel sizes of $\{128, 64, 32, C\}$, where $C$ is the number of channels in the output observation. The first two transposed convolutions use a kernel size of $5 \times 5$, whilst the final two use a kernel size of $6 \times 6$, all with a stride of $2$.

\subsection{Goal-Conditioned Transition Model} The transition dynamics of the GC-RSSM consist of multi-layer perceptrons (MLPs) and a Gated Recurrent Unit (GRU) \cite{cho2014properties}. The recurrent state $\bm{h}_t = f(\bm{h}_{t-1}, \bm{z}_{t-1}, \bm{a}_{t-1})$ is updated using a GRU cell. The input to the GRU is the output of a 1-layer MLP that fuses the previous stochastic latent state $\bm{z}_{t-1}$ and the previous action $\bm{a}_{t-1}$. The prior stochastic state $\tilde{\bm{z}}_t \sim p(\tilde{\bm{z}} \;|\; \bm{h}_t, \bm{g})$ is predicted by concatenating the current deterministic state $\bm{h}_t$ and the goal embedding $\bm{e}_g$. This concatenated vector is passed through an embedding MLP, followed by a state-prior MLP that outputs the mean and variance of a diagonal Gaussian distribution. The standard deviation is parameterised using a Softplus function bounded by a minimum value ($\sigma_{\text{min}} = 0.1$) to prevent numerical instability and mode collapse.

During training, the posterior stochastic state $\hat{\bm{z}}_t \sim q(\hat{\bm{z}} \;|\; \bm{h}_t, \bm{x}_t, \bm{g})$ incorporates the current observation $\bm{x}_t$. The deterministic state $\bm{h}_t$, goal embedding $\bm{e}_g$, and current observation embedding $\bm{e}_t$ are concatenated and processed through an embedding MLP. A subsequent state-posterior MLP outputs the mean and variance of the posterior diagonal Gaussian distribution, similarly bounded by the Softplus function.

\subsection{Reward Predictor} The reward model $r_\theta(\bm{h}_t, \bm{z}_t)$ predicts the expected scalar reward $r_t$ from the current latent state. It takes the concatenated deterministic state $\bm{h}_t$ and stochastic state $\bm{z}_t$, processes them through four hidden layers and projects them to a scalar value via a final linear layer to formulate a 5-layer MLP.

\subsection{Test-Time Planning with the Cross-Entropy Method} \label{sec:planning}

\rev{This appendix completes the CEM procedure of Section \ref{sec:gc-rssm}. To apply the masks, the planner converts each candidate into pixel coordinates via $\bm{p} = (\bm{a}+\bm{1})\cdot H/2$ on the $H\times H$ observation. It clips the survivors to $[-1,1]$ afterwards. Note that $N_e$ is a fixed fraction of the drawn population, not of the survivors. Every survivor therefore becomes an elite whenever fewer than $N_e$ pass the masks. If an iteration leaves none, the planner exits early and executes the current mean. To score a survivor, we roll the goal-conditioned prior out one step and evaluate $r_\theta$ on the predicted latent state. The rejection applies to the first action of a plan only. For $H>1$ we do not know the cloth state after the first pick-and-place, so we score the later actions without mask constraints. We write the executed action as $\bm{a}_t = \operatorname{clip}(\bm{\mu}, -1, 1)$. The simulator's segmentation provides the cloth mask in simulation, and the pipeline of Appendix~\ref{sec:real-world-cloth-seg} provides it in the real world. In the RGB-to-Mask setting, the planner can instead threshold the model's reconstructed observation. The workspace mask marks the ring between the near radius $r_n$ and the far radius $r_f$ around the robot base, in both simulation and the real world (see Appendix~\ref{sec:real-world-setup}). Only the executed action leaves the normalised image space. We map $\bm{a}_t$ back to a pixel of the selected observation window, and then to a pixel of the full camera frame through the crop of Appendix \ref{sec:transfer-heuristic}. We intersect the corresponding camera ray with the table plane, using the camera-to-base transformation of Appendix \ref{sec:calibration}. The resulting pick and place points become two end-effector poses. The robot executes them with the guarded descent and grasp procedure of Appendix \ref{sec:grasping}.}

\section{Diffusion Policy and Data Augmentation}

\subsection{Diffusion Policy for the P\&P Domain} \label{sec:diffusion}

We adopt the Diffusion Policy~\cite{chi2023diffusionpolicy} to generate proximal expert trajectories for garment flattening for offline RL training. We utilise the policy version adapted for the pick-and-place cloth-manipulation domain given by \textsc{DRAPER}~\cite{kadi2025draper}; it formulates quasi-static pick-and-place cloth manipulation as a conditional generative process. Unlike deterministic regressors, the Diffusion Policy learns a stochastic policy that progressively refines noisy proposals into feasible actions conditioned on visual observations.

The Diffusion Policy models the conditional distribution of action sequences $\pi(\bm{A}^t \mid \bm{X}^t)$. It defines the action sequence $\bm{A}^t \equiv \{\bm{a}_i\}_{i=t-T_h}^{t+T_f-1}$ over a prediction horizon $T_p = T_h + T_f$, and the observation history $\bm{X}^t \equiv \{\bm{x}_i\}_{i=t-T_h}^{t}$ over a horizon $T_h$. In our pick-and-place setting, $\bm{a}_i \in \mathbb{R}^d$ represents a continuous vector parameterising grasp locations and target placements with $d=4$.
The framework consists of a forward process and a learned reverse process. The forward process incrementally corrupts the clean action sequence $\bm{A}_0$ with Gaussian noise over $K$ steps. The noisy sequence at step $k$ can be sampled directly as:
\begin{equation}
\bm{A}_k \sim \mathcal{N}\big(\sqrt{\bar{\alpha}_k}\bm{A}_0, (1 - \bar{\alpha}_k)\bm{I}\big),
\end{equation}
where $\bar{\alpha}_k$ is derived from a predefined squared-cosine noise schedule $\{\beta_i\}_{i=1}^k$~\cite{nichol2021improved}.
The reverse process generates actions by iteratively denoising a sample from a standard Gaussian prior, $\bm{A}_K \sim \mathcal{N}(\bm{0}, \bm{I})$. A learned noise prediction network $\epsilon_\theta$ (1D U-Net \cite{ronneberger2015u}) takes the noisy action $\bm{A}_k$, the diffusion step $k$, and a conditioning observation embedding $\bm{e}_x$ as inputs to estimate the injected noise; the embedding $\bm{e}_x$ is derived from a vision encoder (ResNet18 \cite{he2016deep} with Group Normalisation \cite{wu2018group}) applied to $\bm{X}$. For garment flattening tasks, we disable temporal convolution in the 1D U-Net noise predictor by setting $T_h = 0$ and $T_f = 1$, as longer context windows have proved disadvantageous in our experiments.

The refined action sequence for the previous step is computed as:
\begin{equation}
\bm{A}_{k-1} =
\frac{1}{\sqrt{1 - \beta_k}}
\Big(
\bm{A}_k -
\frac{\beta_k}{\sqrt{1 - \bar{\alpha}_k}}
\epsilon_\theta(\bm{A}_k, k, \bm{e}_x)
\Big)
+ \sigma_k \boldsymbol{\epsilon},
\end{equation}
where $\boldsymbol{\epsilon} \sim \mathcal{N}(\bm{0}, \bm{I})$ and $\sigma_k$ controls the stochasticity of the sampler.

Diffusion Policy trains the network to maximise the conditional log-likelihood of the action sequence by minimising the evidence lower bound (ELBO), resulting in the mean squared error (MSE) objective:
\begin{equation}
\mathcal{L} = \mathrm{MSE}\Big(\boldsymbol{\epsilon}_k, \epsilon_\theta(\bm{A}_k, k, \bm{e}_x)\Big).
\label{eqn:diffusion-policy-loss}
\end{equation}
\subsection{Data Augmentation} \label{sec:data-aug}

In order to ensure the robust generalisation of our visual policy, we employ a comprehensive, geometry-aware data augmentation pipeline on the current and goal RGB observations.

The RGB inputs undergo an appearance-level augmentation where they are remapped to a [0, 1] range and injected with subtle Gaussian noise with $\sigma$ of 0.01. Then, we apply rigid spatial transformations to the current state observations to expand the effective dataset and enforce $SE(2)$ equivariance in the learned policy --- we apply continuous random rotations at 1-degree intervals and vertical flips with a 50\% probability to the current RGB and mask images. Crucially, to maintain geometric consistency, we strictly couple these visual transformations with the action space. When the images are rotated or flipped, the corresponding 2D spatial pixel actions (the pick-and-place coordinates) are transformed via equivalent rotation matrices and coordinate inversions, discarding any transformations that push the pixel actions out of bounds.

Finally, to enable robust goal-conditioned behaviour in the real world, the pipeline applies independent affine transformations to the goal RGB observation. These images undergo randomised 2D rigid body transformations, combining rotations and translational shifts (up to a translation scale of 0.2). By independently augmenting the target visual states, we prevent the network from simply memorising exact pixel-to-pixel mappings between the current and goal images, which forces the architecture to learn deeper, spatially robust relational features for the pick-and-place task.

\section{Diffusion Policy vs.\ LaGarNet} \label{sec:comp-diff}

\begin{figure*}[t!]
    \centering
    
    \newlength{\subfigH}\setlength{\subfigH}{0.19\textwidth}
    \subfloat[\centering Validation Trials \label{fig:diffusion-human-val}]{
        \begin{minipage}[t][\subfigH][t]{0.29\textwidth}\centering
        \includegraphics[width=\linewidth]{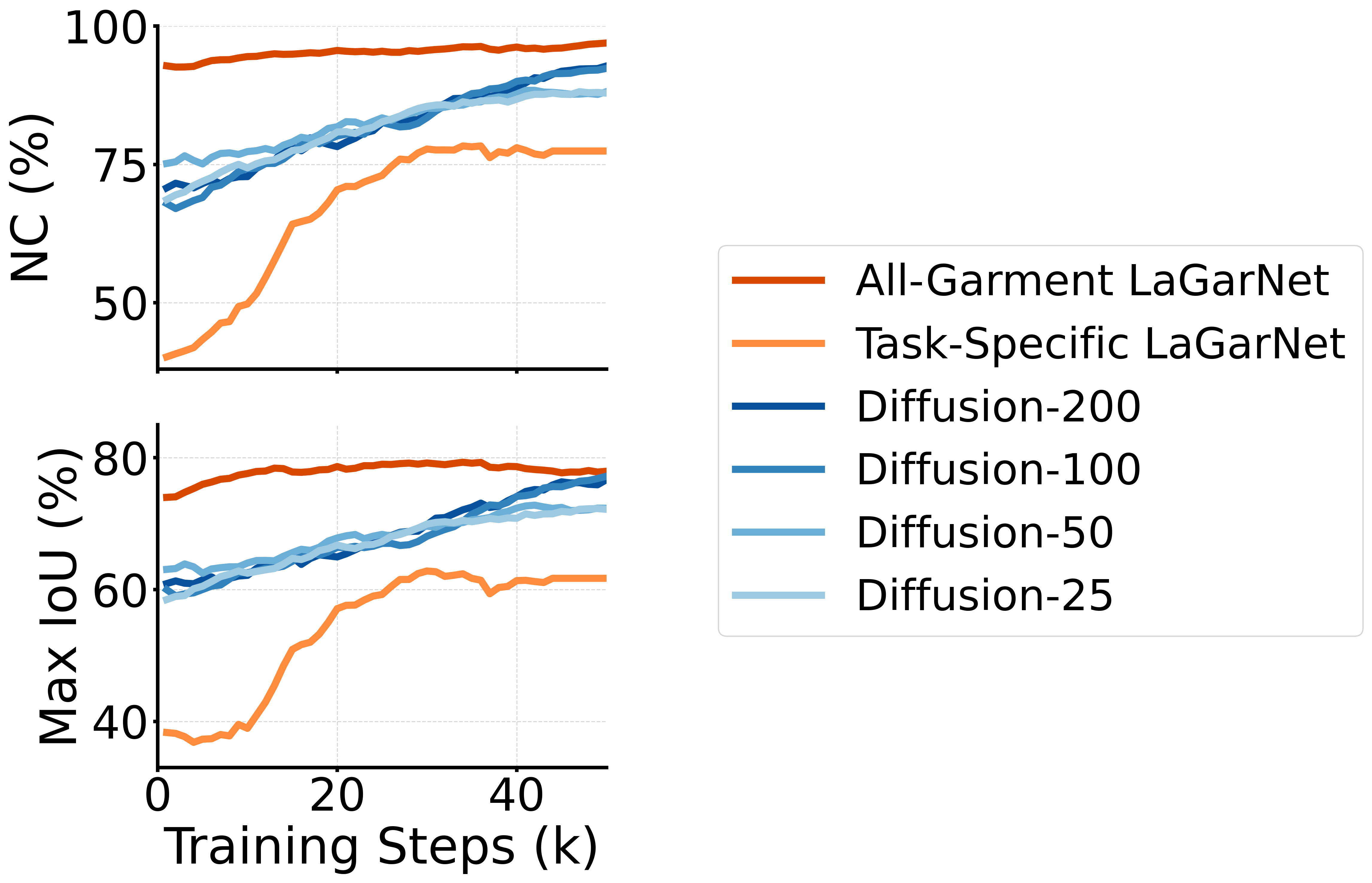}
        \end{minipage}
    }\hfill
    \subfloat[\centering Evaluation Trials \label{fig:diffusion-human-eval}]{
        \begin{minipage}[t][\subfigH][t]{0.185\textwidth}\centering
        \includegraphics[width=\linewidth]{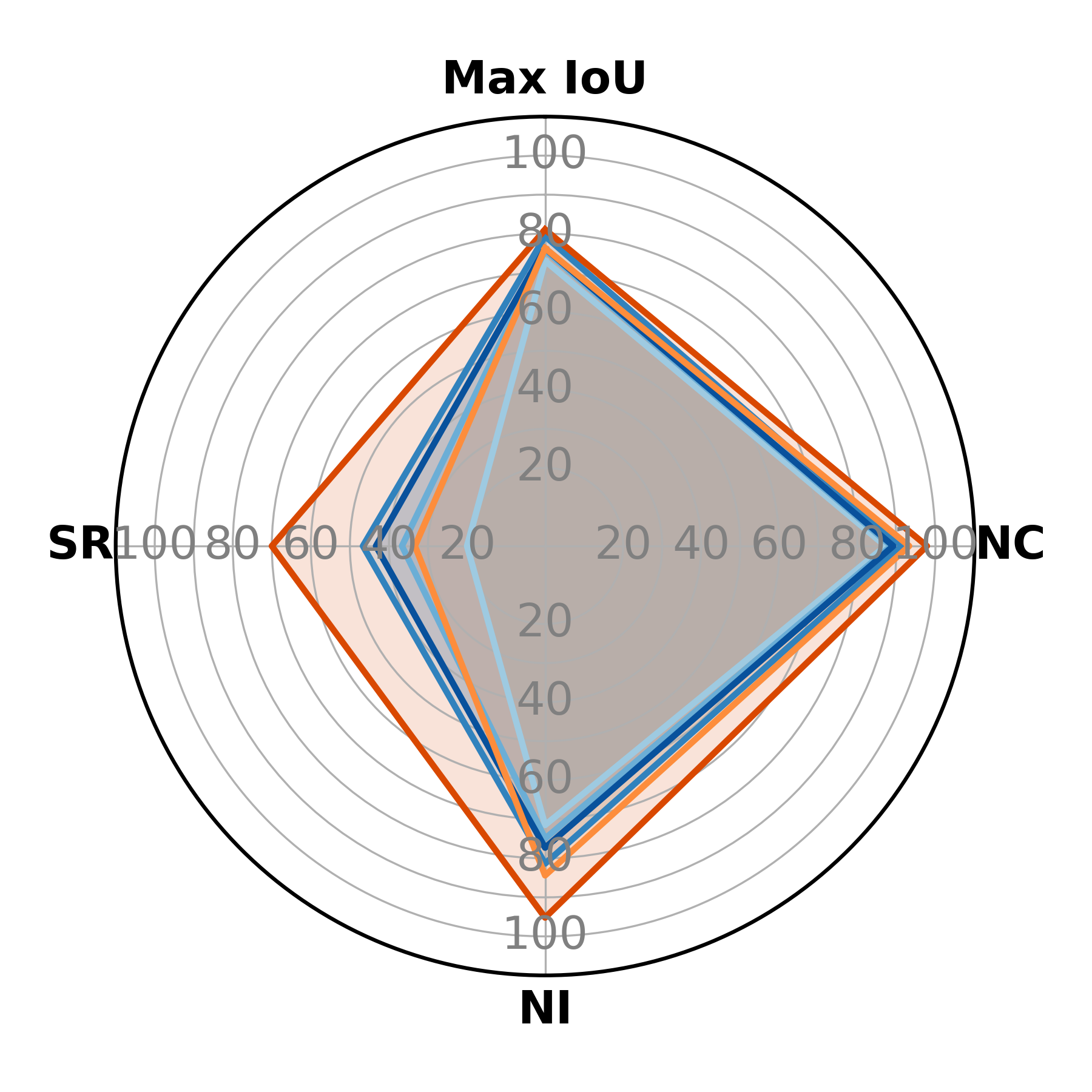}
        \end{minipage}
    }\hfill
    \subfloat[\centering Architectural Ablations \label{tab:comp-diff-lagarnet}]{
        \begin{minipage}[t][\subfigH][t]{0.42\textwidth}\centering
        \resizebox{\linewidth}{!}{
            \begin{tabular}{lcccc}
            \toprule
            Method & SR & Max IoU & NC & NI \\
            \midrule
            GC-DP & 40.0 & 77.6 $\pm$ 6.6 & 89.7 $\pm$ 2.5 & 77.8 $\pm$ 4.8 \\
            GC-DP-Comp & 25.0 & 76.0 $\pm$ 6.6 & 87.2 $\pm$ 1.1 & 72.7 $\pm$ 2.7 \\
            LaGarNet-Comp & 5.0 & 69.1 $\pm$ 2.4 & 77.3 $\pm$ 5.0 & 49.4 $\pm$ 10.9 \\
            LaGarNet-RSSM & 52.5 & 82.0 $\pm$ 3.3 & 92.9 $\pm$ 0.9 & 84.6 $\pm$ 3.0 \\
            
            GC-RSSM-DP & 6.7 & 68.6 $\pm$ 3.4 & 77.4 $\pm$ 2.3 & 50.1 $\pm$ 5.0 \\
            LaGarNet-CNN-DP & 8.3 & 70.0 $\pm$ 3.6 & 78.2 $\pm$ 3.4 & 51.8 $\pm$ 6.5 \\

            \rowcolor{green!20} LaGarNet & 75.0 & 84.5 $\pm$ 3.0 & 95.4 $\pm$ 1.4 & 90.1 $\pm$ 3.4 \\
            \bottomrule
            \end{tabular}
        }
        \end{minipage}
    }
    
    \caption{\rev{Diffusion Policy versus LaGarNet on the simulated longsleeve environment. \textbf{(a)} \resp{Validation performance over 50,000 training steps for task-specific Diffusion Policies trained with varying numbers of human demonstrations. Each panel carries one metric---normalised coverage above, Max IoU below---and every curve is an exponential moving average (EMA, span = 15); the All-Garment and Task-Specific LaGarNet curves are shown for reference.} \textbf{(b)} Evaluation performance of these policies on the four primary metrics. \textbf{(c)} Architectural ablations by action step 20. All values are aggregated over 120 evaluation episodes (30 trials per garment type). The All-Garment LaGarNet trains stably and substantially outperforms every Diffusion Policy variant trained on an equivalent volume of success-only trajectories.}} 
    \label{fig:diff-vs-lagarnet-performance-combined}
\end{figure*}

\rev{Using the setup of Section \ref{sec:simulation}, we ask whether LaGarNet's advantage over the Diffusion Policy stems from explicit forward planning, extended temporal context, or simply data scale. We examine how the number of human demonstrations affects a task-specific Diffusion Policy (Figures \ref{fig:diffusion-human-val} and \ref{fig:diffusion-human-eval}) and compare it against the ablated baselines below.}

\subsection{Baselines} \label{sec:diff-baselines}

\rev{To isolate which architectural components drive LaGarNet's advantage, we design six baselines. (a) \textbf{GC-DP}: a Goal-Conditioned Diffusion Policy taking concatenated current and goal RGB images, trained on 5,000 episodes of diffusion-induced, success-only data spanning all garment types. (b) \textbf{GC-DP-Comp}: GC-DP aligned to LaGarNet's hyperparameters---PlaNet's CNN encoder in place of ResNet-18, the two embeddings concatenated before the downstream networks, and the observation sequence length raised from 1 to 4. (c) \textbf{LaGarNet-Comp}: LaGarNet with its training sequence length cut from 50 to 4 for direct comparison. (d) \textbf{LaGarNet-RSSM}: LaGarNet using its full, unconditioned RSSM dynamics. (e) \textbf{GC-RSSM-DP}: GC-DP-Comp using LaGarNet's frozen pre-trained encoder and latent dynamics, with the noise predictor's conditioning dimension set to 360; each step wipes memory by zeroing the previous latent state and recovers the posterior by feeding a ``no-op'' action into the dynamics model. (f) \textbf{LaGarNet-CNN-DP}: GC-DP-Comp using only LaGarNet's frozen vision encoder, giving a 2048-dimensional embedding of the current and goal images.}

\subsection{Results} \label{sec:diff-result}

\resp{Each baseline isolates one architectural component. Three core components drive the successful performance of LaGarNet over diffusion policy. They are temporal context, goal-conditioned inference, and forward planning. In contrast, visual features and dataset scale does not contribute as much. See Figure \ref{tab:comp-diff-lagarnet} for full comparisons.}

\resp{\emph{Temporal context is critical.} Reducing training sequences to 4 steps degrade performance---SR drops drastically to 5.0\%. The belief state requires sequence history; short histories cannot disambiguate folded garments. In addition, \emph{Goal-conditioned latent inference} matters highly---unconditioned dynamics only reach 52.5\% SR. Goal-conditioning adds 22.5 points of SR. \emph{Forward planning turns coverage into success}---LaGarNet outperforms GC-DP by 35.0 SR points; it also gains 5.7 NC points, achieving through using mixed-quality data. We learned that regression can learn crude cloth spreading, but true success requires predicting action outcomes and forward planning selects actions for these effects.}

\resp{\emph{Alternative explanations do not hold up.} \emph{Visual features do not cause this gap}---using LaGarNet's frozen encoder degrades DP performance. Testing its full latent dynamics performs worse than base GC-DP. Furthermore, \emph{dataset scale is not responsible}---increasing demonstrations to 200 raises validation NC to 93\%. Max IoU also increases to 77\% (Figure \ref{fig:diffusion-human-val}). Yet, all DP variants still underperform LaGarNet. LaGarNet exceeds 93\% NC within 1,000 updates. Ultimately, \emph{test-time search requires a goal-aware latent state.} This state needs deep temporal context. Together, they enable effective test-time search, answering our central research question.}

\section{Latent Reconstruction Quality} \label{sec:recon-appendix}

\begin{figure}[t!]
    \centering
    \includegraphics[width=0.45\textwidth]{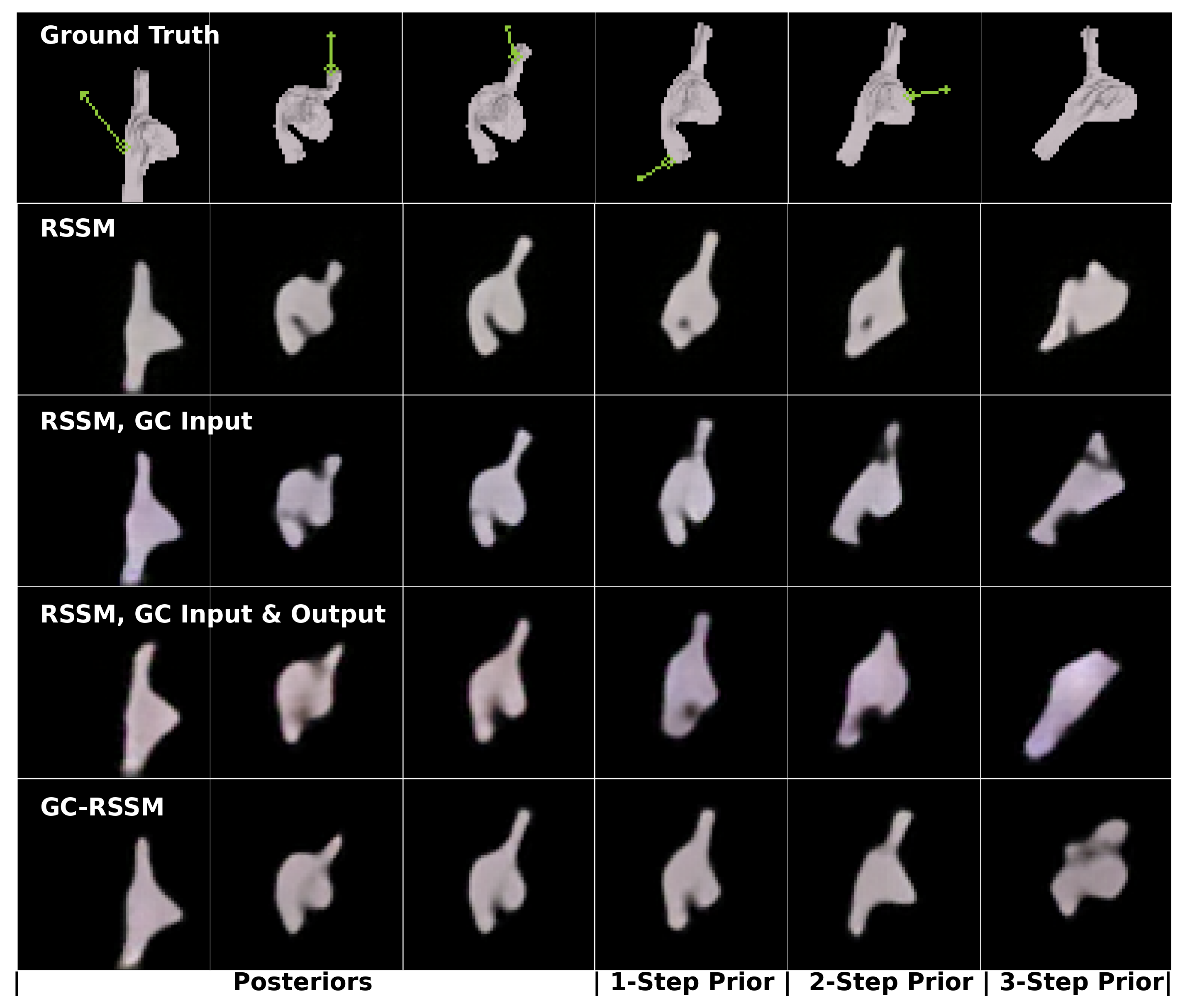}
    \caption{\ral{Observation reconstructions of the latent-dynamics models ablated in Figure \ref{fig:goal-ablation}. Each grid is a $64 \times 64$ RGB image; green arrows in the first row are the pick-and-place actions taken.} GC-RSSM tracks the garment sleeves in the prior steps better than the other RSSM variants.}
    \label{fig:recon-quality}
\end{figure}

\ral{Figure \ref{fig:recon-quality} shows the reconstructions behind the scores of Figure \ref{tab:recon-quality}. The posterior reconstructions are close for every variant, so the models separate only on the priors, where the prediction must be made without the current observation. GC-RSSM keeps the sleeves and the garment outline legible two and three steps ahead, whereas the unconditioned RSSM and RSSM-GC-IO blur them into the body of the garment. This is what the SSIM column of Figure \ref{tab:recon-quality} measures, and it is the representation the planner scores candidate actions against.}

\section{Simulation Garment Distribution} \label{sec:garment-dist}

\rev{Figure \ref{fig:garment-distribution} characterises the 30 ``hard'' initial states used for evaluation, in terms of normalised coverage, maximum IoU against the flattened goal, and garment colour.}

\begin{figure*}[t!]
    \centering
    \begin{subfigure}[t]{0.37\linewidth}
        \vspace{0pt}
        \centering \includegraphics[width=\linewidth]{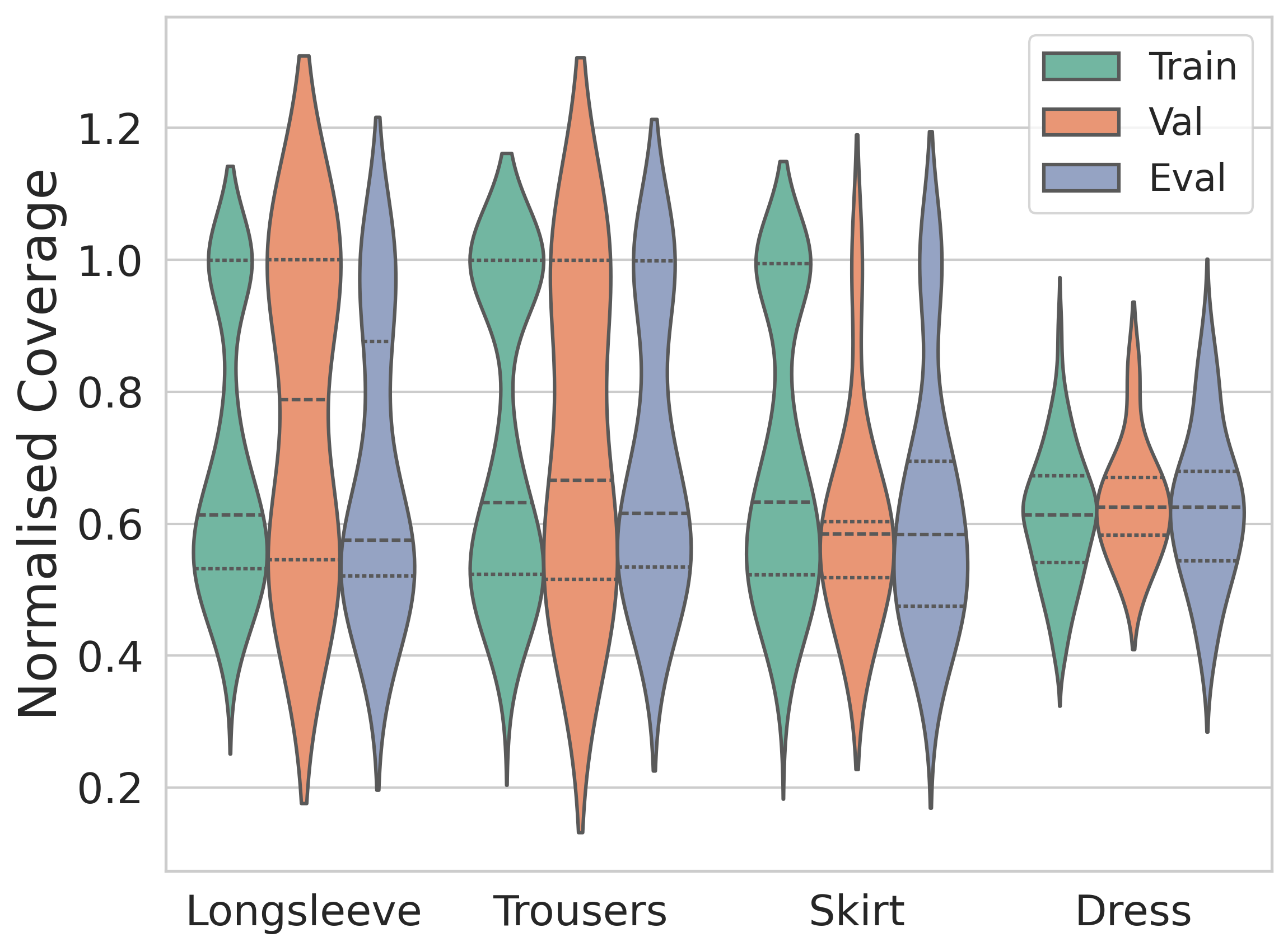}
        \caption{Normalised Coverage Distribution}
        \label{fig:nc-dist}
    \end{subfigure}
    \begin{subfigure}[t]{0.37\linewidth}
        \vspace{0pt}
        \centering \includegraphics[width=\linewidth]{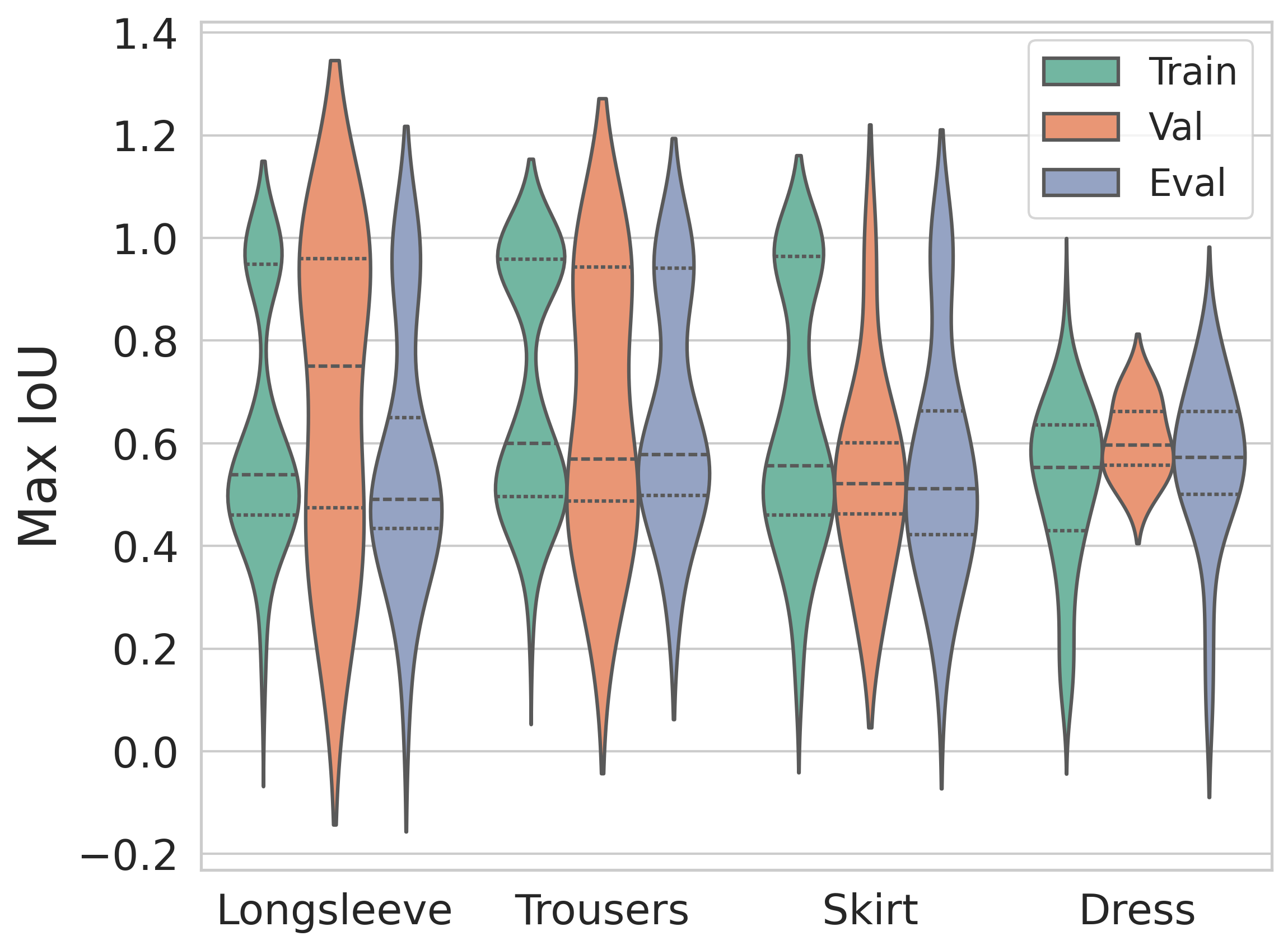}
        \caption{Max IoU Distribution}
        \label{fig:iou-dist}
    \end{subfigure}
    \begin{subfigure}[t]{0.24\linewidth}
        \vspace{0pt}
        \centering \includegraphics[width=\linewidth]{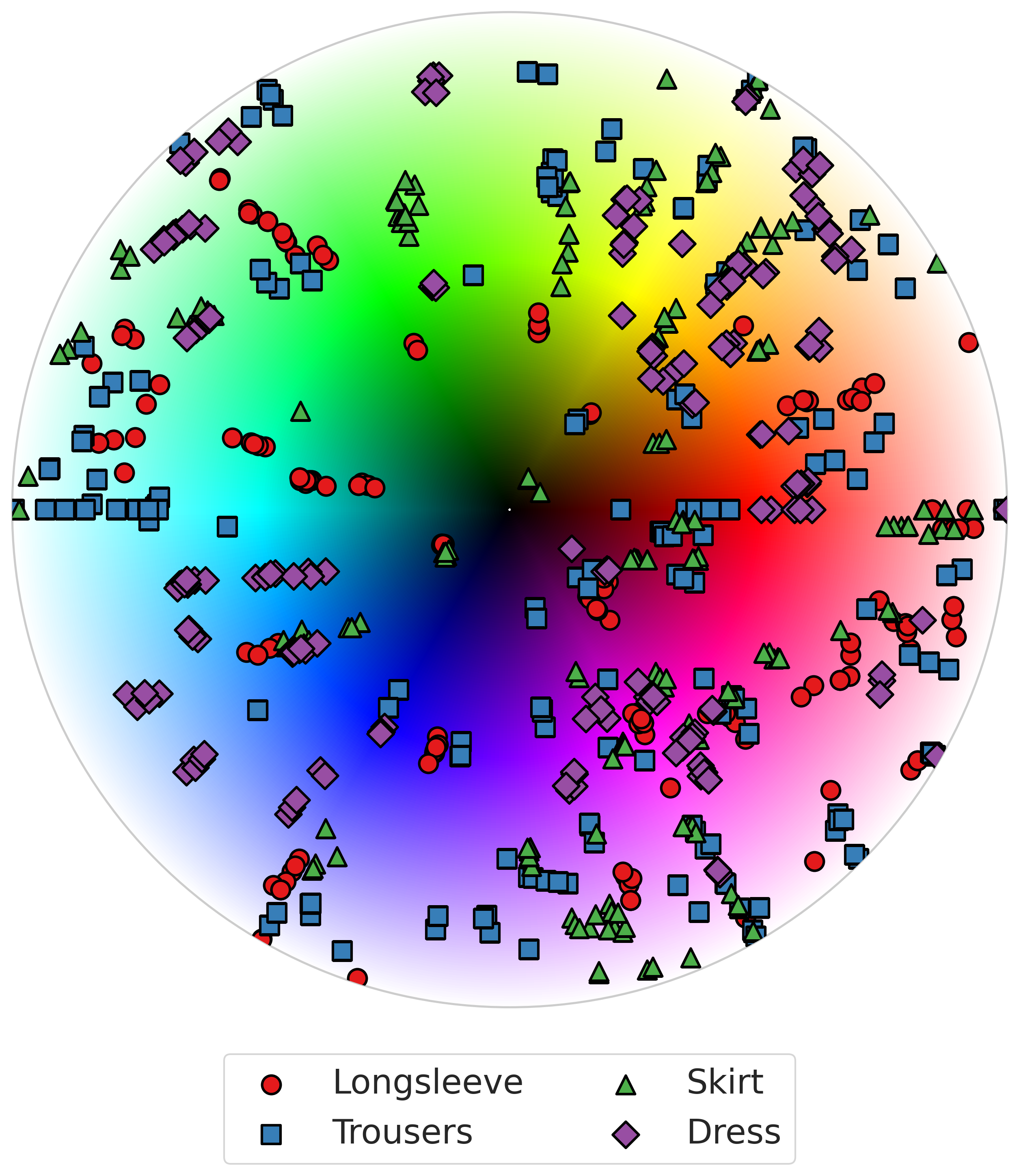}
        \caption{Colour Distribution}
        \label{fig:colour-distribution}
    \end{subfigure}
    \caption{\rev{The initial state distribution of the simulated garments.}}
    \label{fig:garment-distribution}
\end{figure*}

\section{Real Robot Setup} \label{sec:real-world-setup}

In the cloth manipulation literature, robot arms are often fixed on a table \cite{lin2022learning,lee2024learning,canberk2023cloth}, leading to a ring-shaped workspace for Pick-and-Place (PnP) primitives. \rev{Real garments are generally larger than simple fabrics, so it becomes physically impossible to flatten them within a fixed square window inside the workspace of a single robot arm---a common practice in this domain \cite{kadi2025draper, hoque2022visuospatial, lee2021learning}.} This requires us to develop a transfer strategy to bridge the gap between observation and the workspace in addition to the gap between the simulation and the real robot environment.
\begin{figure*}[t!]
    \centering
    
    \begin{subfigure}[b]{0.3\linewidth}
        \centering
        \includegraphics[width=\linewidth]{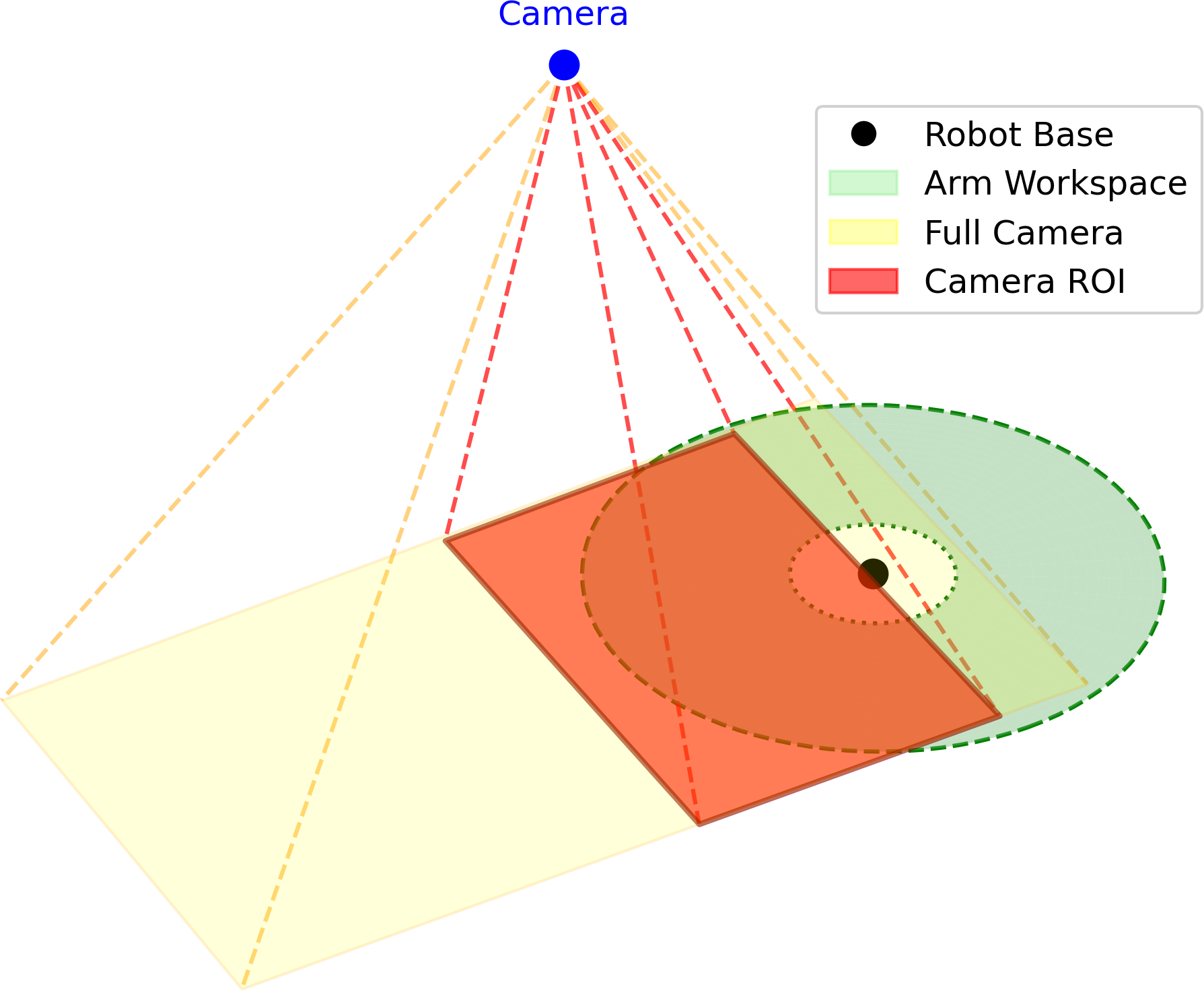}
        \caption{Camera Setup}
        \label{fig:robot-camera}
    \end{subfigure}\hfill
    \begin{subfigure}[b]{0.45\linewidth}
        \centering
        \includegraphics[width=\linewidth]{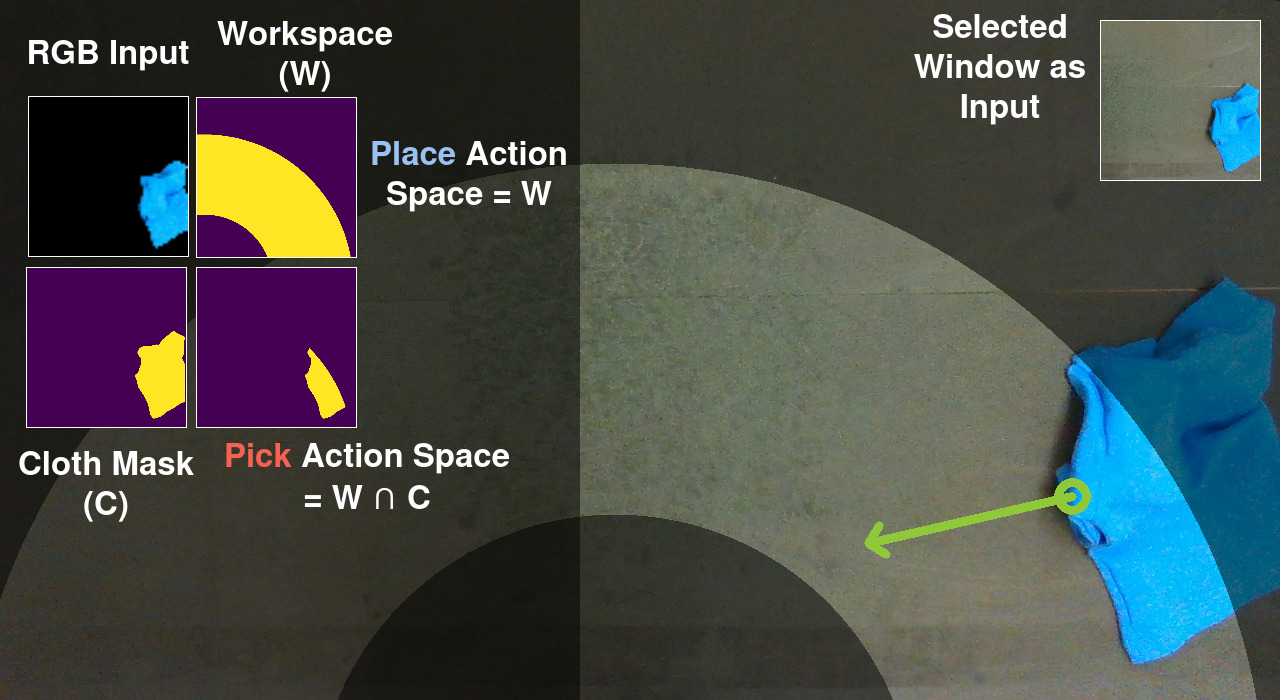}
        \caption{\rev{Heuristic for Transferring the Policy within the Region-of-Interest (ROI)}}
        \label{fig:transfer-heuristic}
    \end{subfigure}
    \begin{subfigure}[b]{0.24\linewidth}
        \centering
        
        \label{fig:setup_c}
        \vspace{0.5em}
        \resizebox{\textwidth}{!}{
        \begin{tabular}{lc}
        \toprule
        \textbf{Camera} & \textbf{Value} \\
        \midrule
        $FoV$ & 69$^\circ$ $\times$ 42$^\circ$\\
        Camera Resolution & 1280 $\times$ 720 px\\
        ROI Resolution & 450 $\times$ 710 px\\
        ROI Start & (700, 10) px \\
        $P_c$ & (0.03, -0.86, 1.66)m \\
        $r_n, r_f$ & 0.2\,m, 0.7\,m\\
        \midrule
        \textbf{Segmentation} & \textbf{Value} \\
        \midrule
        model & \texttt{sam\_vit\_h\_4b8939} \\
        $A_{\text{min}}$  & $1,000$ \\
        $A_{\text{max}}$  & $90,000$ \\
        $\alpha_{border}$ & $0.15$\\
        $S_{\text{min}}$  & $10$ \\
        $V_{\text{min}}$ & $10$ \\
        \bottomrule
        \end{tabular}}
        \caption{Parameters \label{tab:seg-para}}
    \end{subfigure}
    
    \caption{Overview of the real-world deployment setup. \rev{(a) Camera $P_c$ is set at the front of the robot and top of the table surface; whilst the camera can capture the table view highlighted in yellow, our system operates on the Region-of-Interest (ROI) highlighted in red. (b) The heuristic approach for transferring the simulated policy to the real world; the heuristic prioritises the selection of the middle window for network inputs, but if the selected window does not contain any part of the cloth, it re-selects a window that centres around the cloth as shown; the most highlighted part of the rectangular images is the workspace of the \text{UR5e} robot for a particular action step.} (c) The empirical parameters utilised in the segmentation pipeline.}
    \label{fig:real-world-overview}
\end{figure*}
As shown in Figure \ref{fig:ur5e-onrobot}, we employ a UR5e robot arm equipped with an OnRobot RG2 electric parallel gripper to conduct our real-world experiments. We position a RealSense D435i camera $P_c$ in front of the robot and above a table covered with a dark foam surface. As in simulation, we use only RGB frames with automatic exposure as input to our method in the real-world setup.

The workspace of our UR5e robot is defined by a far radius $r_{f}$ and a near radius $r_{n}$. To isolate the effective workspace, we crop the visual input to a Region of Interest (ROI) as in Figure \ref{fig:robot-camera}. Given the top-down camera perspective, we compute the spatial position $p_w$ in the world frame for each pixel $p_x$ in the camera view using the camera intrinsics and the transformation matrix. Consequently, we determine whether $p_x$ lies within the workspace by checking if $r_n \leq ||p_w - b_w||_2 \leq r_f$, where $b_w$ represents the position of the robot base in the world frame. Please refer to Table \ref{tab:seg-para} for the exact parameters of our real-robot setup. \rev{We use the UR-RTDE package \cite{urrtde2025} for conducting Cartesian planning to execute pick-and-place primitives, as it reliably produces stable trajectories.}

This section first introduces the transfer heuristic, which integrates the real robot arm's workspace into LaGarNet's planning method (Section \ref{sec:transfer-heuristic}). Next, we discuss how we conduct our hand-eye calibration (Section \ref{sec:calibration}) for the transformation between the camera pixel space and robot base space. Then, we explain our approach to segmenting the physical garments from the camera observations (Section \ref{sec:real-world-cloth-seg}). Finally, we detail our grasping implementation, designed to ensure the precise and robust picking of garments (Section \ref{sec:grasping}).

\subsection{Transfer Heuristic for Robot's Ring-Shaped Workspace} \label{sec:transfer-heuristic}
Many robot manipulators face workspace constraints, such as limited reach as well as undesirable trajectories caused by inverse kinematics and joint discontinuities. These factors often make parts of the workspace unusable for smooth pick-and-place (PnP) execution. To address this general class of problems, we propose a heuristic workspace sampling method that adapts trained policies to feasible regions of a robot's workspace without retraining. We demonstrate this method on a UR5e robot arm, whose PnP workspace resembles a ring due to both reach limitations and unstable trajectories near the base; the ring-like workspace is defined by a far radius $r_f$ and a near radius $r_n$ on the table surface.

Our heuristic workspace sampling method (see Figure \ref{fig:transfer-heuristic}) builds an observation bridge between the UR5e environment and the policy's training environment. This method prioritises the central square window of the raw RGB input. If the central window does not contain any part of the cloth, it re-selects a window centred on the target cloth. The \text{LaGarNet} latent dynamics network receives observations filtered by the cloth mask in the selected window, and the model predictive control planner samples pick actions at the intersection of the cloth mask and workspace mask, and place actions within the workspace mask in the selected window. Using these two processes, we ensure that \text{LaGarNet} can be transferred to the entire visible workspace of a UR5e robot from a top-down camera for flattening real garments. Note that the dynamic cropping mechanism of the transfer heuristic is only activated at the beginning of an episode if the garment is absent from the central field of view.

\subsection{Hand-Eye Calibration} \label{sec:calibration}

To guarantee precise physical manipulation based on visual input, we conducted a rigorous eye-to-hand calibration to map the Intel RealSense camera's pixel space to the Universal Robots (UR) base frame. Our calibration target is a $5 \times 7$ ChArUco board \cite{garrido2014automatic} featuring 40~mm squares and 30.0~mm ArUco markers (Dictionary 4x4\_50) grasped by the robot's end-effector. To ensure an optimal and mathematically robust solution, we program the robot to traverse a trajectory of 50 carefully designed joint configurations. These poses explicitly sample the $SE(3)$ workspace, incorporating isolated Z-axis depth variations, lateral translations, and decoupled pitch, yaw, and roll rotations. This diverse pose generation prevents singularities and ensures the focal parameters are well-constrained. For each spatial sample, we estimate the target-to-camera transformation utilising a Perspective-n-Point solver based on the interpolated ChArUco corners. Finally, we compute the fixed camera-to-base transformation matrix by solving the standard $AX=XB$ formulation using the Park and Martin method \cite{park1994robot}.

\subsection{Real-World Garment Segmentation} \label{sec:real-world-cloth-seg}
Following standard practices in the domain \cite{canberk2023cloth, lin2022learning, kadi2025draper}, we utilise the Segment Anything Model (SAM) \cite{kirillov2023segment} to isolate the target garment. This segmentation is crucial for stripping the background from policy inputs and enabling accurate real-world evaluation. Given an RGB observation, SAM generates a set of $K$ binary candidate masks. To efficiently isolate the true garment, we convert the image to the HSV colour space and apply a multi-stage heuristic filtering pipeline. First, we apply spatial pruning: masks are retained only if their total pixel area $A \in [A_{\text{min}}, A_{\text{max}}]$. To eliminate broad background surfaces, we reject any candidate whose active border pixels exceed a fraction $\alpha_{border}$ of the image perimeter, formalised as $P_{\text{mask}} > \alpha_{border} \times 2(h + w)$. Next, we apply chromatic and variance filters to distinguish the garment from its environment. For each mask, we compute the mean saturation $\bar{S}$ and RGB intensity variance $V$. \rev{We discard low-saturation grey areas, for which $\bar{S} < S_{\text{min}}$.} We further reject grey backgrounds by ensuring $ V \geq V_{min}$. Finally, the optimal mask is selected by maximising a utility score that prioritises large, distinctly coloured objects $\text{Score}_i = A_i \cdot \bar{V}_i$. This pipeline reliably yields high-fidelity garment segmentations for downstream manipulation tasks. The specific parameters used in our real-world experiments are detailed in Table \ref{tab:seg-para}.
\subsection{Grasping} \label{sec:grasping}

Accurately grasping garments without failure remains a significant challenge for parallel grippers. We build upon the grasping protocol proposed by DRAPER \cite{kadi2025draper}, which adjusts the pick position on an eroded mask of the garment and ensures the axis connecting the two gripper fingers is perpendicular to the target mask cutline. However, while DRAPER utilises a custom pincher extension, our approach employs a standard OnRobot RG2 parallel gripper. To mitigate misgrasping errors, we initialise the gripper to a 3~cm opening width before the approach towards garments. The end-effector then executes a guarded descent, lowering gradually until the robot's force-torque sensor detects an upward reaction force of approximately 5~N. This force-based threshold ensures the garment is firmly contacted while preventing gripper faults or misgrasps caused by excessive collision resistance from the underlying workspace table.

\section{Real-World Failure Cases} \label{sec:failure-appendix}
    
\begin{figure}[t!]
    \centering
    \includegraphics[width=\columnwidth]{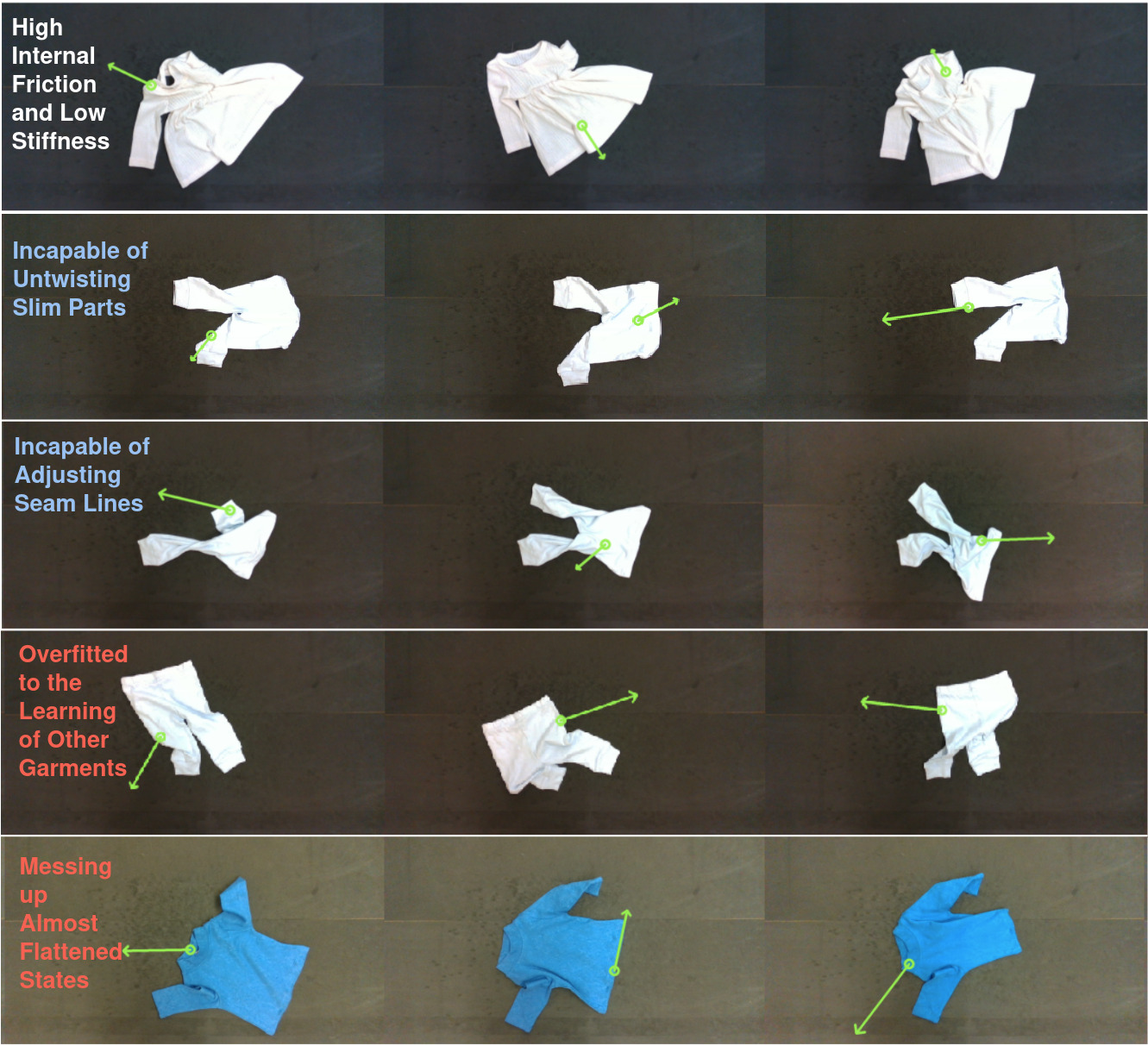}
    \caption{\ral{Visual failure cases of LaGarNet in the real world. Text colour indicates the source of the failure: white for mechanical limitations, blue for oversights in our assumptions, and red for the learning process.}}
    \label{fig:failure-visuals}
\end{figure}

\ral{Figure \ref{fig:failure-visuals} shows the failure modes counted in Table \ref{tab:failure-rates}. The mechanical cases follow from the single-arm parallel gripper, which cannot hold one part of the garment whilst pulling another and so leaves stubborn wrinkles in place. The assumption cases arise where the setup itself is the limit, such as the policy being unable to tell the front of a garment from its back. The learning cases are the ones the model could in principle avoid: disturbing an already flattened garment, or treating a trouser leg as though it were a sleeve.}

\renewcommand{\refname}{Appendix References}

\fi

\end{document}